\documentclass[nohyperref]{article}

\usepackage{microtype}
\usepackage{graphicx}
\usepackage{subfigure}
\usepackage{booktabs} 

\usepackage{hyperref}

\usepackage[accepted]{icml2022}

\usepackage{amsmath}
\usepackage{amssymb}
\usepackage{mathtools}
\usepackage{amsthm}

\usepackage[capitalize,noabbrev]{cleveref}

\theoremstyle{plain}
\newtheorem{theorem}{Theorem}

\theoremstyle{definition}

\theoremstyle{remark}

\usepackage[textsize=tiny]{todonotes}
\usepackage[utf8]{inputenc}
\usepackage{natbib}
\usepackage[acronym,smallcaps,nowarn]{glossaries}
\glsdisablehyper
\makeglossaries
\usepackage{color}

\newcommand{\opt}{*}
\newcommand{\dopt}{d^\opt}

\newcommand{\mathbold}[1]{\ensuremath{\boldsymbol{\mathbf{#1}}}}

\newcommand{\g}{\,|\,}

\newcommand{\nestedmathbold}[1]{{\mathbold{#1}}}

\newcommand{\mbphi}{\nestedmathbold{\phi}}

\DeclareMathOperator*{\argmax}{arg\,max}

\newcommand{\cA}{\mathcal{A}}
\newcommand{\cD}{\mathcal{D}}

\newcommand{\cM}{\mathcal{M}}
\newcommand{\cN}{\mathcal{N}}
\newcommand{\cP}{\mathcal{P}}

\newcommand{\cR}{\mathcal{R}}
\newcommand{\cS}{\mathcal{S}}
\newcommand{\cT}{\mathcal{T}}
\newcommand{\cU}{\mathcal{U}}

\newcommand{\E}{\mathbb{E}}

\newcommand{\bbR}{\mathbb{R}}
\newcommand{\bbN}{\mathbb{N}}

\newacronym{DL}{dl}{deep learning}
\newacronym{VI}{vi}{variational inference}
\newacronym{BO}{bo}{Bayesian optimization}
\newacronym{DOE}{doe}{design of experiments}
\newacronym{SGD}{sgd}{stochastic gradient descent}
\newacronym{EIG}{eig}{expected information gain}
\newacronym{BOED}{boed}{Bayesian optimal experimental design}
\newacronym[longplural=Markov decision processes]{MDP}{mdp}{Markov decision process}
\newacronym{RL}{rl}{reinforcement learning}
\newacronym{naiverl}{naive-rl}{naive reinforcement learning}
\newacronym{PCE}{pce}{prior contrastive estimation}
\newacronym{VPCE}{vpce}{variational prior contrastive estimation}
\newacronym{SMC}{smc}{sequential Monte Carlo}
\newacronym{SPCE}{{s}pce}{sequential prior contrastive estimation}
\newacronym{DAD}{dad}{deep adaptive design}
\newacronym{SED}{sed}{sequential experimental design}
\newacronym{REDQ}{redq}{Randomized ensembled double q-learning}
\newacronym{SNMC}{snmc}{sequential nested monte carlo}
\newacronym{CES}{ces}{constant elasticity of substitution}
\newacronym{hipmdp}{hip-mdp}{Hidden Parameter MDP}

\icmltitlerunning{Optimizing Sequential Experimental Design with Deep Reinforcement Learning}

\begin{document}

\twocolumn[
\icmltitle{Optimizing Sequential Experimental Design with Deep Reinforcement Learning}



\icmlsetsymbol{equal}{*}

\begin{icmlauthorlist}
\icmlauthor{Tom Blau}{data}
\icmlauthor{Edwin V. Bonilla}{data}
\icmlauthor{Iadine Chades}{lw}
\icmlauthor{Amir Dezfouli}{data}
\end{icmlauthorlist}

\icmlaffiliation{data}{CSIRO's Data61, Australia}
\icmlaffiliation{lw}{CSIRO's Land and Water, Australia}

\icmlcorrespondingauthor{Tom, Blau}{tom.blau@data61.csiro.au}

\icmlkeywords{Machine Learning, ICML}

\vskip 0.3in
]

\printAffiliationsAndNotice{}

\begin{abstract}
    Bayesian approaches developed to solve the optimal design of sequential experiments are mathematically elegant but computationally challenging. Recently, techniques using amortization have been proposed  to make these Bayesian approaches practical, by training a parameterized policy that proposes designs efficiently at deployment time. However, these methods may not sufficiently explore the design space, require access to a differentiable probabilistic model and can only optimize over continuous design spaces. Here, we address these limitations by showing that the problem of optimizing policies can be reduced to solving a Markov decision process (MDP). We solve the equivalent MDP with modern deep reinforcement learning techniques. Our experiments show that our approach is also computationally efficient at deployment time and exhibits state-of-the-art performance on both continuous and discrete design spaces, even when the probabilistic model is a black box. 
\end{abstract}

\section{Introduction}
One of the fundamental building blocks of scientific investigation is the development of predictive models that describe natural phenomena. Such models must be learned from data, which can only be acquired by conducting experiments. Since experiments are often costly and time-consuming, this naturally leads to the question of how to design experiments so that they are maximally informative. The formalism of \acrlong{BOED} \citep[\acrshort{BOED},][]{lindley1956measure} casts this problem as optimization of expected information gain. \acrshort{BOED} is a popular approach in many scientific fields~\cite{chaloner1995bayesian,ryan-doe-review-2016}.

Following a \gls{BOED} approach, let $p(y \g \theta, d)$ be a probabilistic model describing some phenomenon. Here $d$ represents our independent variables, or experimental \textit{design}, $y$ is the dependent variables, or experimental \textit{outcome}, and $\theta$ the parameters of the model. We do not know the exact values of $\theta$, but we have some prior belief about it represented as a distribution $p(\theta)$. 
The goal is then to find the design 
$\dopt$ that maximizes the \gls{EIG} from the experiment:
\begin{equation}
    d^* = \argmax_{d \in \mathcal{D}} \E_{p(y \g d)} \left[H(p(\theta)) - H(p(\theta \g y, d ))\right],
   \label{eq:ent_eig}
\end{equation}
where $p(y \g d)$ can be computed as an expectation over the original model, $H(\cdot)$ is the entropy of a distribution, and $\cD$ is the space of possible designs. Simply put, this means maximizing the expected decrease in entropy from the prior to the posterior. In the sequential setting, after conducting an experiment and observing its outcome, the new posterior $p(\theta \g y, d)$ is used instead of the prior and a new design is computed. This can be repeated for as long as resources allow, a process called \textit{iterated experimental design}.

Evaluating the above expression requires computing an expectation w.r.t.~the joint probability distribution $p(y, \theta \g d) = p(y \g d) p(\theta \g y, d)$, which can be challenging by itself~\cite{hong2009estimating}. An even more difficult problem occurs when we have a budget of $T$ experiments and wish to find the \textit{non-myopically} optimal design 
that maximizes expected information gain w.r.t.~all future experiments. This requires computing repeatedly nested expectations, with the depth of nesting growing at a rate of $O(T)$, and the convergence rate of evaluating this expectation being $O(n^{-\frac{1}{T+1}})$ ~\cite{rainforth2018nesting}, where $n$ is the number of samples. Approaches to non-myopically optimal design that rely on estimating this expectation are therefore intractable.

Recently amortized methods have been introduced to solve the problem of non-myopic optimization~\cite{foster2021deep,ivanova2021implicit}. In this amortized approach a \textit{design policy} is introduced that maps the history of designs and outcomes to the next design. A lower bound estimate of the non-myopic \gls{EIG} is optimized w.r.t.~the parameters of this policy, rather than optimizing the designs directly. After training, the policy achieves inference times orders of magnitude faster and \gls{EIG} superior to (non-amortized) iterative, myopically optimal methods. However, the approach involves backpropagation through a policy network, and hence requires that the probabilistic model be differentiable and the design space continuous. This makes these types of approaches inapplicable in problems where the design space is discrete, or where gradients of the probabilistic model are not available.

To address these shortcomings, we propose to formulate the \gls{SED} problem as a \gls{MDP}~\citep{bellman1957markovian}, and learn the design policy using \gls{RL}. Since our \gls{RL} agents exploit the Policy Gradient Theorem~\cite{sutton1999policy}, there is no need for the \gls{EIG} objective to be differentiable. Furthermore, \gls{RL} algorithms exist for optimizing agents in discrete decision spaces. Finally, modern \gls{RL} techniques have many features such as stochastic policies and entropy-based regularization terms that help improve exploration of the solution space. We show that our approach, using deep \gls{RL} methods, achieves state-of-the-art \gls{EIG} in both continuous and discrete experimental design problems, even when the probabilistic model is a black-box, i.e. its gradients are not available. Further, we show that correct formulation of the \gls{MDP} is critical to the success of our method, and naive conversions of \glspl{SED} to \glspl{MDP} are significantly less performant.

\section{Background}
In the \gls{BOED} framework, given a probabilistic model $p(y \g \theta, d)$ and a prior distribution over the parameters $p(\theta)$ we aim to find a design $\dopt$ that maximizes an expected utility, where the expectation is taken over the joint distribution $p(y, \theta \g d)$. It is a basic result in \gls{BOED} that choosing the utility $\cU = \log p(\theta \g y, d) - \log p(\theta)$, leads to maximizing the \gls{EIG} objective:
\begin{align}
    EIG(d)
    &= \E_{p(y \g d)} \left[H(p(\theta)) - H(p(\theta \g y, d ))\right].
\end{align}
Hence, solving a \gls{BOED} problem maximizing utility $\cU$  is equivalent to optimizing the $\gls{EIG}$ w.r.t.~the design variables $d$. In general, the \gls{EIG} cannot be evaluated in closed form, and must be estimated numerically.~\citet{foster2020unified} introduced the \gls{PCE} lower bound:
\begin{equation}
    PCE(d, L) \equiv \E_{p(\theta_{0:L})p(y\g \theta_0, d)} \left[ \log \frac{p(y \g \theta_0, d)}{\frac{1}{L+1} \sum^L_{l=0} p(y \g \theta_ l, d)} \right]
    \label{eq:pce}
\end{equation}
where $\theta_{0:L}$ are sampled i.i.d. from the prior $p(\theta)$ and $\theta_{1:L}$ are \textit{contrastive samples} that re-weight the experiment outcome under alternative realizations of $\theta$. With this estimator it is simple to optimize the design of a single experiment in isolation.

\subsection{Sequential Experimental Design}

The naive approach to \gls{SED} is to iteratively optimize the \gls{PCE} bound, conduct the experiment, update the posterior using Bayes' rule, and repeat. Let $h_t = (d_{1:t}, y_{1:t})$ be the history of experimental designs and outcomes at time $t$. At each iteration, the prior $p(\theta)$ is replaced by an intermediate posterior $p(\theta \g h_t)$.

This approach has two significant drawbacks. The first is that computing posteriors is expensive in the general case, and here one must be computed at each time step. The second is that the approach is myopic: it only optimizes the immediate \gls{EIG} from the next experiment, without taking into consideration how this would affect future experiments.

Suppose that rather than optimizing the designs directly, we instead optimize a policy $\pi : \mathcal{H} \rightarrow \mathcal{D}$ which maps the experimental history at time $t$ to the next design, i.e., $d_t = \pi(h_{t-1})$. The expected total information gain of following this policy is~\cite{foster2021deep}:
\begin{align}
    & EIG_T(\pi)  \equiv
        \E_{p(\theta)p(h_T \g \theta, \pi)} \left[
        \log
        \frac{p(h_T \g \theta, \pi)}{p(h_T \g \pi)} \right], \label{eq:total-eig} \\
    & p(h_T \g \theta, \pi) =   
        \prod_{t=1}^T p(y_t \g \theta, d_t) \label{eq:full-likelihood},
\end{align}
where $p(h_T \g \pi) = \E_{p(\theta)} p(h_T\g \theta, \pi)$ is the marginal likelihood of the history. Note that the total \gls{EIG} above does not require expectations w.r.t.~the posterior over model parameters and,
instead, the expectations are w.r.t.~the prior $p(\theta)$. 
\Cref{eq:total-eig} can be estimated using the \gls{SPCE} lower bound: 
\begin{align}
    \begin{split}
        sPCE(\pi, L, T) &\equiv \quad\E_{p(\theta_{0:L}) p(h_T \g \theta_0, \pi)} \left[ g(\theta, h_T) \right] \\
        g(\theta, h_T) &= \log \frac{p(h_T \g \theta_0, \pi)}{\frac{1}{L+1} \sum^L_{l=0} p(h_T \g \theta_l, \pi)}, 
    \end{split}   \label{eq:spce}
\end{align}
where the required expectations can be estimated numerically by sampling $\theta_{0:L}$ i.i.d.~from $p(\theta)$ and $h_T$ from \cref{eq:full-likelihood}.
A parameterized policy $\pi_{\mbphi}(\cdot)$, given for example by a neural network, can then be found by maximizing~\cref{eq:spce} using  stochastic gradient ascent. 
This method of optimizing the policy w.r.t.~the \gls{SPCE} objective is known as \gls{DAD} \cite{foster2021deep}.

\gls{DAD} avoids the estimation of posteriors altogether and amortizes the cost at training time via the parameterized policy $\pi_\mbphi$, thus achieving fast computation of designs at deployment time. However, as we shall see in the next section, we can leverage the \gls{SPCE} bound above to propose a  more general approach that can explore the design space more efficiently, and handle discrete designs and black-box probabilistic models. We will do this by reducing the policy optimization problem to that of solving a \gls{MDP}.

\subsection{Reinforcement Learning}
\glsreset{MDP}
Reinforcement learning is based on the \gls{MDP} framework. An \gls{MDP} is a tuple of the form $<\mathcal{S}, \mathcal{A}, \mathcal{T}, \mathcal{R}, \rho_0>$ where $\mathcal{S}$ is the space of possible system states; $\mathcal{A}$ is the space of possible control actions, $\mathcal{T}$ is the transition dynamics $\mathcal{T} : \mathcal{S} \times \mathcal{A} \rightarrow \cP_s(\cS)$, giving for each state-action pair a distribution over states; 
$\mathcal{R}$ is a reward function $\mathcal{R} : \mathcal{S} \times \mathcal{A} \times \mathcal{S} \rightarrow \bbR$, giving the expected immediate reward gained for taking an action under a current state and ending up in another state; and $\rho_0$ is the distribution of initial states. 
In \gls{RL} we are interested in policies $\pi: \cS \rightarrow  \cP_a(\cA)$ that map the current state to a distribution over actions. 
Our goal is then to solve the \gls{MDP}, i.e., to find an optimal policy $\pi^*$ that maximizes the expected discounted return
\begin{equation}
    J(\pi) \equiv \E_{\mathcal{T}, \pi, \rho_0} \left[ \sum^T_{t=1} \gamma^{t-1} \mathcal{R}(s_{t-1}, a_{t-1}, s_t 
    )\right], \label{eq:return}
\end{equation}
with \textit{discount factor} $\gamma \in [0,1]$ and \textit{time horizon} $T \in \bbN$.
\section{RL for Sequential Experiment Design}
In order to optimize \gls{SED} problems with \gls{RL} algorithms, we must first formulate them as \glspl{MDP} that satisfy the equality
\begin{equation}
    J(\pi) = sPCE(\pi, L, T).
\end{equation}
The naive approach is to assign the posterior as the state $s_t = p(\theta \g h_t)$, the actions as designs, and the difference of entropies as the reward $r_t = H(s_{t-1}) - H(s_t)$. Unfortunately, this formulation would require the inference of a posterior in each timestep. We can avoid this expensive repeated inference by using $g(\theta, h_t)$ from~\cref{eq:spce} as our reward instead. This implies that the state must include the true value of $\theta$ as it is an input to $g(\theta, h_t)$. This is troublesome in the \gls{MDP} framework since $\theta$ is not observable at test time, and is not affected by actions. Thus, we resort to the \gls{hipmdp} framework for our formulation.

\subsection{Hidden Parameter MDPs}

A \gls{hipmdp}~\cite{doshi2016hidden} is a tuple $<\cS, \cA, \Theta, \cT, \cR, \gamma, \rho_0, P_\Theta>$ where $\cS, \cA$ are the same as in a traditional \gls{MDP} but we add a parameter space $\Theta$ that parameterizes the reward and transition functions, meaning $r_t = \cR(s_{t-1,} a_{t-1}, s_t, \theta)$ and $s_t \sim \cT(s_t \g s_{t-1}, a_{t-1}, \theta)$. The realization $\theta$ is drawn once at the start of an episode from the prior $P_\Theta$ and is fixed until the end of the episode. Note the overlap in notation with $\theta$ from our probabilistic model is intentional, as both terms refer to the same variables. The expected return becomes:
\begin{equation}
    J(\pi) \equiv \E_{\mathcal{T}, \pi, \rho_0, P_\Theta} \left[ \sum^T_{t=1} \gamma^{t-1} \mathcal{R}(s_{t-1}, a_{t-1}, s_t, \theta)\right].
\end{equation}
This leads to a formulation of \gls{SED} as a \gls{hipmdp} where $P_\Theta = p(\theta)$, states are histories, actions are designs, and the reward and transition functions are $g(\cdot, \cdot)$ and the likelihood model, parameterised by $\theta$. However, we cannot simply set $r_t = g(\theta, h_t)$ because this would count the contribution of the $t^{th}$ experiment $T-t$ times. The naive solution is to set $r_t = 0\ \forall t < T$ and compute $g(\cdot, \cdot)$ only for the terminal reward.~\Cref{thm:naive_equivalence} establishes that the solution to such a \gls{hipmdp} also optimizes the corresponding \gls{SED} problem.

\begin{theorem}\label{thm:naive_equivalence}
    Let $\cM$ be a \gls{hipmdp} where $s_t = h_t; a_{t-1} = d_t \ \forall t \in [1,T]$ and initial state distribution and transition dynamics are $\rho_0 = h_0 \sim \delta(\emptyset); P_\Theta = \theta_{0:L} \sim p(\theta)$ and $\cT = p(y_t \g h_{t-1}, d_t, \theta_0)$ respectively. If $\gamma = 1$ and the reward function is:
    \begin{align}
        \cR(s_{t-1}, a_{t-1}, s_t, \theta) =
        \begin{cases}
            0,& \text{if } t < T \\
            g(\theta, h_t),& \text{if } t = T
        \end{cases}
    \end{align}
    then the expected return satisfies:
    \begin{equation}
        J(\pi) = sPCE(\pi, L, T).
    \end{equation}
\end{theorem}

For proof cf.~\cref{proof:naive_equivalence}. We are therefore free to use any \gls{RL} algorithm of our choosing to learn a policy that will optimize the return, and hence the \gls{SPCE}.

\subsection{The SED MDP}
While the above treatment is sufficient for optimizing \gls{SED} problems with \gls{RL} algorithms, it still has a few shortcomings: the state is not truly Markovian as it contains the entire history, and the reward signal is sparse, which is generally known to hinder performance in \gls{RL} settings. In this section we will take a closer look at the state and reward definition, in order to formulate a more precise \gls{hipmdp} that addresses these problems. We turn first to the issue of the Markov property.

Compactly representing the history is a well-studied problem in partially observable MDPs. A common approach is to learn such representations from data~\cite{gregor2019shaping}. Consider the permutation invariant representation proposed by~\citet{foster2021deep}:
\begin{equation}
    B_{\psi,t}\equiv \sum_{k=1}^t ENC_\psi(d_k, y_k),
\end{equation}
where $ENC_\psi$ is an encoder network with parameters $\psi$. Although not originally intended to do so, this history summary induces a Markovian structure. Since it can be decomposed as $B_{\psi,t} = B_{\psi,t-1} + ENC_\psi(d_t, y_t)$, we do not need the whole history $h_t$ to compute the next summary $B_{\psi,t+1}$. Rather, it is sufficient to store the most recent summary $B_{\psi,t}$. Note that at training time we will still need to keep entire histories in the replay buffer in order to learn $\psi$.

The history summary suffices as an input for the policy, but rewards depend on $g(\theta, h_t)$, which cannot be computed from $B_{\psi,t}$. To address both the issues of the Markov property and of sparsity, we propose the following reward function, which estimates the marginal contribution of the $t^{th}$ experiment to the cumulative EIG:
\begin{align}
    \begin{split}
        \mathcal{R}(s_{t-1}, a_{t-1}, s_t, \theta) =& \log p(y_t \g \theta_0, d_t) \\
        &- \log (C_t \cdot \mathbf{1}) + \log (C_{t-1} \cdot \mathbf{1}),
    \end{split} \label{eq:marginal_reward}\\
  \text{where }  C_t =& \left[ \prod^t_{k=1} p(y_k \g \theta_l, d_k) \right]_{l=0}^L 
\end{align}
is a vector of history likelihoods, such that each element $c_{t,l}$ is the likelihood of observing the history $h_t$ if the true model parameters were $\theta_l$. We use $\mathbf{1}$ to denote a vector of ones of the appropriate length.

Our proposed reward has several elegant properties: first, it is in general nonzero for all timesteps, providing a dense learning signal for the \gls{RL} agent. Second, the reward function assigns each experiment its immediate contribution to the final EIG. Therefore the sum of rewards over an entire trajectory (or any prefix of a trajectory) is exactly $g(\theta, h_t)$ from~\cref{eq:spce} for that trajectory (or for the prefix). In the special case of $t=1$, the reward is exactly the non-sequential \gls{PCE}. Third, we do not need to keep the entire history in memory in order to compute the reward---it is sufficient to store the most recent history likelihood $C_t$ and experiment outcome $y_t$. This is because the history likelihood decomposes as $C_t = C_{t-1} \odot \left[ p(y_t \g \theta_l, d_t) \right]_{l=0}^L$, where $\odot$ denotes the Hadamard (or elementwise) product.

Note that the above formulation means we need to store different history representations for the policy and the reward function, and the state is thus a concatenation of both. The transition dynamics involve sampling the outcome of an experiment from the model, and updating the history summaries $B_\psi$ and history likelihoods $C$ accordingly:
\begin{align}
    \begin{split}
        y_t &\sim p(y_t \g d_t, \theta_0) , \\
        B_{\psi,t} &= B_{\psi,t-1} + ENC_\psi(d_t, y_t) \text{ and}\\
        C_t &= C_{t-1} \odot \left[ p(y_t \g \theta_l, d_t) \right]_{l=0}^L . \label{eq:dynamics}
    \end{split}
\end{align}
Putting it all together, we have the following \gls{hipmdp}:
\begin{itemize}
    \item $\cS$: the current experiment outcome $y_t$ and the history summaries and likelihoods used by $\pi$ and $R$, respectively. $s_t = (B_{\psi,t}, C_t, y_t)\ \forall t\in[0,T]$.
    \item $\cA$: is the design space with  $a_{t-1} = d_t \ \forall t\in[1,T]$.
    \item $R$: the reward function of~\cref{eq:marginal_reward}.
    \item $\cT$: the transition dynamics of~\cref{eq:dynamics}.
    \item $\rho_0$: the initial history is always the empty set, thus $\rho_0 = (B_{\psi,0}, C_0, y_0) = (\mathbf{0}, \mathbf{1}, \emptyset)$.
    \item $\Theta$: is the space of model parameters and $P_\Theta = p(\theta)$.
    \item $\pi$: at time $t$, a policy $\pi(a_t \g s_{t})$  maps $B_{\psi,t}$ to a distribution over designs $d_{t+1}$. Therefore, unlike \citet{foster2021deep}, our policy is stochastic.  
\end{itemize}

\Cref{thm:equivalence} establishes that the solution to the above \gls{hipmdp} also optimizes the corresponding \gls{SED} problem:
\begin{theorem}\label{thm:equivalence}
    Let $\mathcal{M}$ be a \gls{hipmdp} where $s_t = (B_{\psi,t}, C_t, y_t); a_{t-1} = d_t \ \forall t \in [1,T]$ and initial state distribution is $\rho_0 = (\mathbf{0}, \mathbf{1}, \emptyset); P_\Theta = \theta_{0:L} \sim p(\theta)$ and reward and transition functions follow~\cref{eq:marginal_reward,eq:dynamics}, respectively. If $\gamma = 1$ then the expected return satisfies:
    \begin{equation}
        J(\pi) = sPCE(\pi, L, T). \label{eq:general_equivalence}
    \end{equation}
\end{theorem}

\begin{proof}
    Here we will provide a sketch of proof. For the full proof, see~\cref{proof:equivalence}. The proof relies on showing that the terms on both sides of~\cref{eq:general_equivalence} are equal given the conditions of the theorem. As each side is an expectation, we show that the distributions on both sides are identical, and that the terms inside the expectation are equal for any realisation of the random variables.
\end{proof}
\subsection{Advantages of the RL Formulation} 
Armed with this recipe for converting \gls{SED} problems into \glspl{MDP}, we can now deploy \gls{RL} algorithms of our choosing to learn design policies off-line, and then use the policies for rapid on-line experiment design.

\subsubsection{Discrete-Design Spaces and Black-Box Likelihoods}
Learning design policies in this way has several advantages over the \gls{DAD} approach. First, \gls{DAD} is incapable of handling problems with discrete design spaces, a limitation explicitly acknowledged by~\citet{foster2021deep}. In contrast, there are many \gls{RL} algorithms that can efficiently solve \glspl{MDP} with discrete action spaces. Another advantage of our method is that \gls{RL} algorithms do not need the reward function to be differentiable, whereas \gls{DAD} must have access to the gradients of the \gls{SPCE} (w.r.t.~policy parameters) in order to optimize the design policy. Hence, in cases where the likelihood is available but is not differentiable w.r.t.~the design (for example, if the probabilistic model contains an argmax), \gls{RL} can be used but \gls{DAD} cannot.

\subsubsection{Exploration vs Exploitation}
Finally, the trade-off between exploration and exploitation is a significant and long-standing challenge in reinforcement learning~\cite{sutton2018reinforcement,weng2020exploration} as well as in experiment design~\cite{robbins1952some}. At any point in time, an agent must choose either to take an action that is optimal w.r.t.~the expected return given current knowledge (exploitation), or an action that will lead to better knowledge and thus better returns in the future (exploration).
Modern \gls{RL} algorithms have many features that help improve exploration, such as random noise in the policy, regularization terms, and so forth \cite{ladosz2022exploration}. \gls{DAD}, lacking these features, is a pure exploitation algorithm. Consequently, it may explore the design space insufficiently and thus get trapped in a poor local optimum. While some exploration features can perhaps be incorporated into the \gls{DAD} framework, it would require significant engineering and scientific effort to do so. On the other hand, high-quality implementations of these features are already available in the \gls{RL} framework, at no additional cost to practitioners.

\section{Experimental Results}

In this section we compare our approach to several baselines in order to empirically establish its theoretical benefits. We will examine experimental design problems with both continuous and discrete design spaces, as well as the case where the likelihood model is not differentiable. Our method\footnote{See \url{https://github.com/csiro-mlai/RL-BOED} for source code.}, which we term the \gls{RL} approach, is implemented in the Pyro~\cite{bingham2018pyro} and Garage~\cite{garage} frameworks. We used the \gls{REDQ} \cite{chen2020randomized} algorithm to train all of our policies. As an ablation, we include a version of our approach based on~\cref{thm:naive_equivalence}, denoted as \gls{naiverl}. This ablation shares the permutation-invariant policy of the full \gls{RL} approach, so that it evaluates only the effect of our original reward formulation and not the effect of the policy architecture we inherit from~\citet{foster2021deep}.

\subsection{Continuous Design Space}

We begin with an examination of continuous design problems. This is the only scenario in which the \gls{DAD} baseline is applicable. Additionally, we have the \gls{VPCE} baseline, which involves iteratively optimizing the myopic \gls{PCE} bound by gradient descent w.r.t.~the next design, then estimating a posterior with variational inference~\cite{foster2020unified}. Finally, a random policy is included to provide a naive baseline.

\subsubsection{Source Location Finding}
The first problem we consider is \textit{source location}, previously studied by~\citet{foster2021deep}. In this problem there are $2$ signal sources positioned in $2$-dimensional space. The strength of the signal decays with distance from its respective sources, and the signals of the different sources combine additively. The signal can be measured with some observation noise at arbitrary points in the plane, and the design is the $2$-d coordinates at which to take this sample. For full details see~\cref{app:source}.

\begin{figure}[ht]
\vskip 0.1in
\begin{center}
\centerline{\includegraphics[width=\columnwidth]{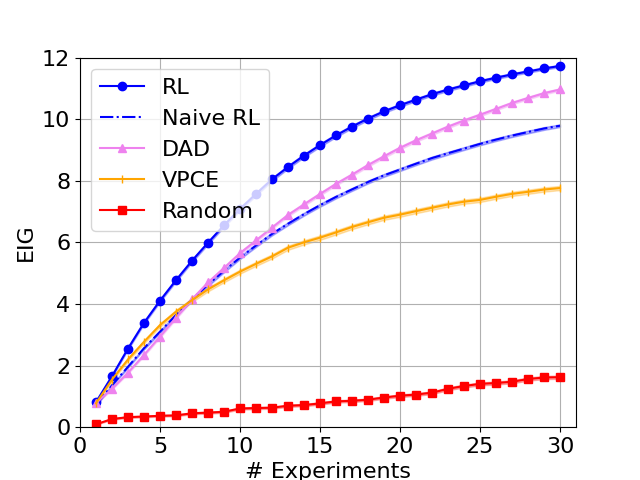}}
\caption{\gls{EIG} for the source location problem, estimated using \gls{SPCE} with $L=1e6$. Trendlines are means and shaded regions are standard errors aggregated from $2000$ rollouts (\gls{RL} and \gls{DAD}) or $1000$ rollouts (\gls{VPCE} and random).}
\label{fig:source}
\end{center}
\vskip -0.2in
\end{figure}

\cref{fig:source} shows the expected information gain of the various methods for an experimental budget of $30$ experiments, where \gls{EIG} was estimated using the \gls{SPCE} lower bound with $1e6$ contrastive samples. Trendlines are the means and shaded regions are the standard errors aggregated from $2000$ rollouts of \gls{RL} and \gls{DAD} or $1000$ rollouts for \gls{VPCE} and random. The standard errors are very small due to the large number of rollouts. Since \gls{RL} is notorious for being highly sensitive to the random seed, we split the $2000$ rollouts of \gls{RL} among $10$ agents each trained with its own random seed.

The \gls{RL} method outperforms all baselines by a considerable margin, and the effect size is several times the magnitude of the standard error. Interestingly, \gls{DAD} underperforms the myopic methods in the first few experiments before overtaking them. This is unsurprising, as \gls{DAD} prioritizes high \gls{EIG} at $t=30$ at the expense of \gls{EIG} at any timestep $t < 30$. What is notable is that the \gls{RL} method, which similarly prioritizes long-term over short-term gain, is nonetheless able to achieve better \gls{EIG} than the other baselines at every timestep. This may be due to the superior exploration capabilities that \gls{RL} provides.

\cref{table:source} focuses on the \gls{EIG} at $t=30$, including both lower and upper bound estimates. The upper bound is based on the \gls{SNMC} estimator~\cite{foster2021deep}. These results emphasize the significance of the gap between \gls{RL} and the baselines. The improvement in \gls{RL}'s lower bound over \gls{DAD}'s is an order of magnitude bigger than the standard error, and the improvement over the non-amortized methods is even bigger. In terms of the upper bound, \gls{RL} performs comparably to \gls{DAD} and significantly better than all other baselines. It is important to note, however, that both amortized methods are trained to maximize the lower bound, not the upper bound.

\begin{table}[t]
\caption{Lower and upper bounds on \gls{EIG} at $t=30$ for the source location problem, computed using \gls{SPCE} and \gls{SNMC} with $L=1e6$ respectively. Means and standard errors from $2000$ (\gls{RL} and \gls{DAD}) or $1000$ (\gls{VPCE} and random) rollouts.}  
\label{table:source}
\vskip 0.15in
\begin{center}
\begin{small}
\begin{sc}
\begin{tabular}{lrr}
\toprule
Method & Lower bound & Upper bound \\
\midrule
Random    & 1.624$\pm$0.053 & 1.639$\pm$0.057  \\
VPCE    & 7.766$\pm$0.069 & 7.802$\pm$0.072  \\
DAD    & 10.965$\pm$0.041 & 12.380$\pm$0.086  \\
\textbf{RL}    & \textbf{11.73$\pm$0.040} & 12.362$\pm$0.062  \\
Naive RL    & 9.789$\pm$0.045 & 9.898$\pm$0.049  \\
\bottomrule
\end{tabular}
\end{sc}
\end{small}
\end{center}
\vskip -0.1in
\end{table}
\subsubsection{Constant Elasticity of Substitution}
The next problem we examine is the \gls{CES}, previously studied by~\citet{foster2020unified}. This is a behavioral economics problem in which a participant compares $2$ baskets of goods and rates the subjective difference in utility between the baskets on a sliding scale from $0$ to $1$. A probabilistic model with latent variables $(\rho, \alpha, u)$ represents the response function, and the goal is to design pairs of baskets in order to infer the value of the latent variables. Each basket consists of $3$ constrained values, so that the design space is $6$-dimensional. For full details see~\cref{app:ces}.

\begin{figure}[ht]
\vskip 0.1in
\begin{center}
\centerline{\includegraphics[width=\columnwidth]{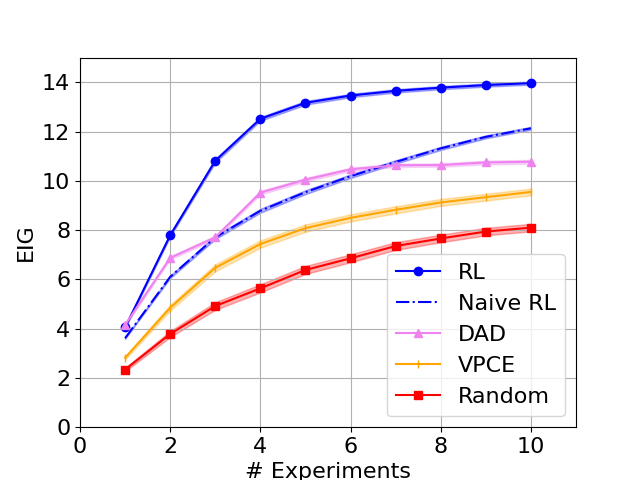}}
\caption{\gls{EIG} for the \gls{CES} problem, estimated using \gls{SPCE} with $L=1e7$. Trendlines are means and shaded regions are standard errors aggregated from $2000$ rollouts (\gls{RL} and \gls{DAD}) or $1000$ rollouts (\gls{VPCE} and random).}
\label{fig:ces}
\end{center}
\vskip -0.2in
\end{figure}

In~\cref{fig:ces} we see the \gls{EIG} for our proposed \gls{RL} method and the baselines over the course of $10$ experiments. \gls{EIG} was estimated using \gls{SPCE} with $1e7$ contrastive samples, and again we show means and standard errors. As in the source location case, the \gls{RL} approach outperforms all baselines, and both amortized methods outperform the non-amortized, myopic methods. However, a notable difference is that in the \gls{CES} problem the performance of \gls{DAD} begins to converge much earlier in the sequence of experiments than \gls{RL} does. This is likely to be an exploration issue, which arises from the nature of the \gls{CES} problem. The experiment outcome $y$ is the result of a censored sigmoid function, and as a consequence of this, much of the design space maps to an outcome on the boundary of the observation space. This leads to a challenging exploration problem, as the surface of the objective function has many suboptimal local maxima in which to be trapped. The \gls{RL} algorithm has a stochastic policy with Gaussian noise that diminishes as performance improves, and this allows it to explore the design space more fully and find better maxima. The \gls{DAD} algorithm, on the other hand, uses a deterministic policy and struggles to explore regions of design space far from where that policy was initialized.

\begin{figure}[ht]
\vskip 0.1in
\begin{center}
\centerline{\includegraphics[width=\columnwidth]{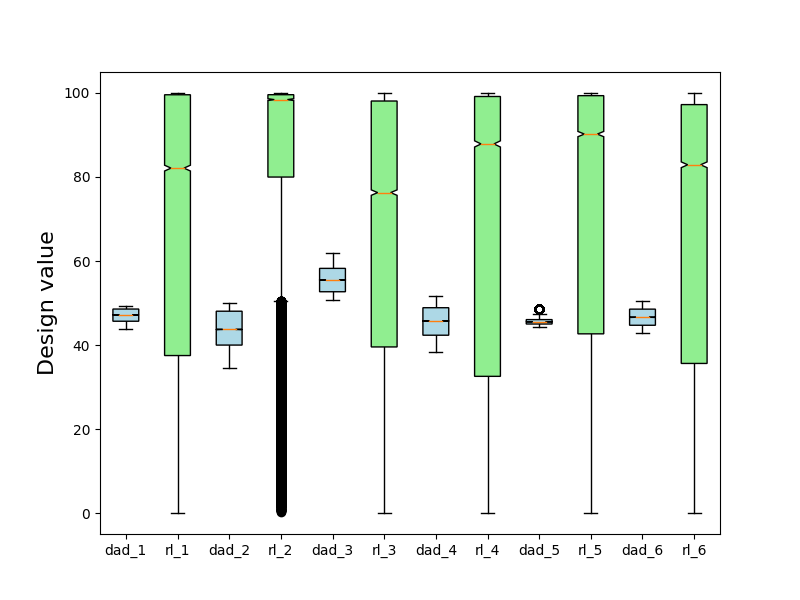}}
\caption{Distributions of designs proposed by trained \gls{RL} (green boxes) and \gls{DAD} (blue boxes) policies for the \gls{CES} problem. One box plot for each of the $6$ elements of the design vector. Data collected from $2000$ rollouts. Boxes indicate the interquartile range, with a notch at the median. Whiskers extend up to $1.5$ times the interquartile range.}
\label{fig:dist}
\end{center}
\vskip -0.2in
\end{figure}

To illustrate the exploratory behavior of both algorithms, we plot the distributions over designs proposed by the trained policies of both methods. As can be seen in~\cref{fig:dist}, the \gls{DAD} policies only propose designs in a narrow range of the design space, relatively near to its center. Consequently, for some samples of $\theta \sim p(\theta)$ the \gls{DAD} policies will not be able to propose effective designs, which limits their \gls{EIG}. The \gls{RL} policies, on the other hand, are capable of proposing designs from all over the space. Because the \gls{RL} algorithm has good exploration, it encountered situations in training where diverse designs have lead to improved outcomes. Consequently, the learned policies propose such designs when needed.

~\cref{table:ces} shows the performance advantage of the \gls{RL} approach even more clearly than~\cref{fig:ces}. At $t=10$, the \textit{lower bound} of \gls{EIG} for \gls{RL} is greater than the \textit{upper bound} for \gls{DAD}, and this difference is several times greater than the standard errors. The gap between \gls{RL} and the myopic baselines is even bigger.

\begin{table}[t]
\caption{Lower and upper bounds on \gls{EIG} at $t=10$ for the \gls{CES} problem, computed using \gls{SPCE} and \gls{SNMC} with $L=1e7$ respectively. Means and standard errors from $2000$ (\gls{RL} and \gls{DAD}) or $1000$ (\gls{VPCE} and random) rollouts.}
\label{table:ces}
\vskip 0.15in
\begin{center}
\begin{small}
\begin{sc}
\begin{tabular}{lrr}
\toprule
Method & Lower bound & Upper bound \\
\midrule
Random    & 8.099$\pm$0.153 & 16.451$\pm$0.685  \\
VPCE    & 9.547$\pm$0.137 & 24.396$\pm$2.024  \\
DAD    & 10.774$\pm$0.077 & 13.374$\pm$0.150  \\
\textbf{RL}    & \textbf{13.965$\pm$0.064} & \textbf{17.794$\pm$0.226}  \\
Naive RL    & 12.131$\pm$0.058 & 15.641$\pm$0.166  \\
\bottomrule
\end{tabular}
\end{sc}
\end{small}
\end{center}
\vskip -0.1in
\end{table}

\subsection{Discrete Design Space}

We now turn to an examination of experimental design problems where the design space is discrete. As mentioned previously, \gls{DAD} is inapplicable in this situation. The \gls{VPCE} baseline also needs to be modified: instead of using gradient descent to optimize the (non-sequential) \gls{PCE} estimator, we compute it for every possible design and take the argmax. Note that this modification is tractable because the \gls{VPCE} baseline is myopic, and we only need to compute it for $|\cA|$ designs. A similar adjustment to \gls{DAD} is not tractable, as we would need to compute the \gls{SPCE} for $|\cA| ^ T$ designs.

We consider a prey population problem, where we control the initial population of a prey species and measure the number of individuals that were consumed by predators after a period of $24$ hours. The goal is to estimate the \textit{attack rate} and \textit{handling time} of the predator species. For full details see~\cref{app:prey}. This setting was previously studied by~\citet{moffat2020sequential}, and we include their \gls{SMC} method as a baseline.

\begin{figure}[ht]
\vskip 0.1in
\begin{center}
\centerline{\includegraphics[width=\columnwidth]{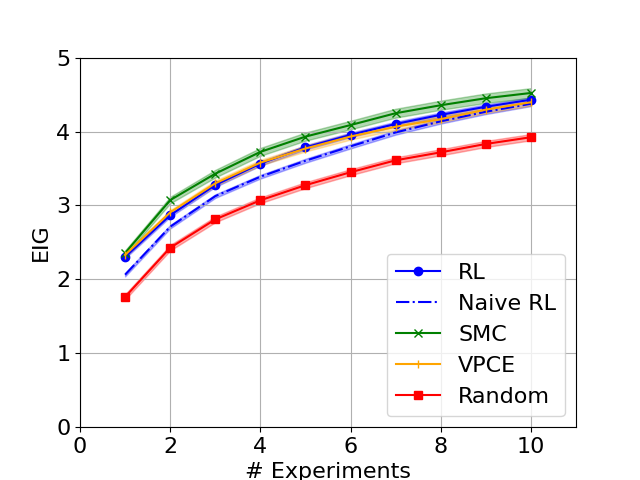}}
\caption{\gls{EIG} for the prey population problem, estimated using \gls{SPCE} with $L=1e6$. Trendlines are means and shaded regions are standard errors aggregated from $2000$ rollouts (\gls{RL}),  $1000$ rollouts (\gls{VPCE} and random) or $500$ rollouts (\gls{SMC}).}
\label{fig:prey}
\end{center}
\vskip -0.2in
\end{figure}

\gls{EIG} for this problem, given a budget of $10$ experiments, is shown in~\cref{fig:prey}. All $3$ methods outperform the naive random baseline. The \gls{RL} method achieves similar results to the \gls{SMC} and \gls{VPCE} baselines. \gls{SMC} has a small performance advantage, but the difference is within the error bars.  This is further emphasized by~\cref{table:prey}, which lists the lower and upper bound estimates for each method at $t=10$. The gap in the lower bound between \gls{SMC} and \gls{RL} is smaller than the standard error, in spite of the large number of replicates. It is possible that with additional replication a statistically significant difference would emerge, but the relative effect size would still be on the order of $1\%$.

It is noteworthy that the \gls{RL} method does not outperform the myopic baselines, since at least in theory it has the capability to find non-myopic solutions. There are a number of possible explanations for this result.
The first is that the myopic baselines compute an explicit posterior after each experiment, and as such have a more accurate basis to perform each optimization than the \gls{RL} method, which must implicitly encapsulate this information in its policy and critic networks. It is possible that the state representation being learned by the \gls{RL} agent is degrading its overall performance, and that better representation learning would lead to improvement over the myopic baselines. Another possibility is that the \textit{optimality gap} between the myopically optimal solution and the non-myopically optimal solution is small for the prey population problem. It is impossible to know for sure if this is the case as no theoretical bounds are known for either type of optimal solution.

\begin{table}[ht]
\caption{Lower and upper bounds on \gls{EIG} at $t=10$ for the prey population problem, computed using \gls{SPCE} and \gls{SNMC} with $L=1e6$ respectively. Means and standard errors from $2000$ rollouts (\gls{RL}),  $1000$ rollouts (\gls{VPCE} and random) or $500$ rollouts (\gls{SMC}).}
\label{table:prey}
\vskip 0.15in
\begin{center}
\begin{small}
\begin{sc}
\begin{tabular}{lrr}
\toprule
Method & Lower bound & Upper bound \\
\midrule
Random    & 3.923$\pm$0.042 & 3.925$\pm$0.043  \\
VPCE    & 4.396$\pm$0.046 & 4.42$\pm$0.050  \\
SMC    & 4.521$\pm$0.065 & $4.523\pm$0.063  \\
RL    & 4.456$\pm$0.032 & 4.459$\pm$0.033  \\
Naive RL    & 4.375$\pm$0.032 & 4.376$\pm$0.032  \\
\bottomrule
\end{tabular}
\end{sc}
\end{small}
\end{center}
\vskip -0.1in
\end{table}

\subsection{Computational Benefits of the RL Appproach at Deployment}
Ultimately, \gls{RL} can achieve comparable \gls{EIG} to state-of-the-art methods which compute explicit posteriors, and require considerably more computation time at deployment.~\cref{table:time} shows the time in seconds that each method needs to compute a single proposed design. For the prey population problem, the trained \gls{RL} agents propose designs orders of magnitude faster than the \gls{SMC} and \gls{VPCE} baselines. Indeed, the \gls{RL} agents are close to the deployment time of the random baseline, which is so fast and trivial that no algorithm can be reasonably expected to beat it. In the context of continuous design spaces, \gls{RL} achieves comparable deployment time to \gls{DAD}, in the sense that the deployment time of both is imperceptible to humans. This is unsurprising, as both approaches compute the design by means of a forward pass through a neural network. \gls{RL} is slightly slower because it must also sample from a probability distribution.

\begin{table}[t]
\caption{Deployment time in seconds by method and problem. Means and standard errors computed from $100$ replications.}
\label{table:time}
\vskip 0.15in
\begin{center}
\begin{small}
\begin{sc}
\begin{tabular}{llr}
\toprule
Problem & Method & Deployment time (s)\\
\midrule
CES  &   Random    & 2.37e-5$\pm$1.51e-7   \\
&   VPCE    & 146.944$\pm$1.397  \\
&   PCE-BO    & 23.830$\pm$0.771  \\
&   DAD    & 1.25e-4$\pm$3.37e-7  \\
&   RL    & 1.35e-3$\pm$4.18e-6  \\
\midrule
Prey population  &   Random    & 1.29e-4$\pm$4.94e-7   \\
&   VPCE    & 20.550$\pm$1.893  \\
&   SMC    & 81.252$\pm$4.310  \\
&   RL    & 1.50e-3$\pm$3.23e-6  \\
\bottomrule
\end{tabular}
\end{sc}
\end{small}
\end{center}
\vskip -0.1in
\end{table}

\subsection{Non-differentiable Likelihood}

One of the claimed benefits of the \gls{RL} approach is that it can learn design policies even without access to gradients of the likelihood. To demonstrate this, we re-run the \gls{CES} problem with a pytorch \textit{no\textunderscore grad} context around the likelihood function. This means that gradients were not computed for the likelihood model.

\begin{figure}[ht]
\vskip 0.1in
\begin{center}
\centerline{\includegraphics[width=\columnwidth]{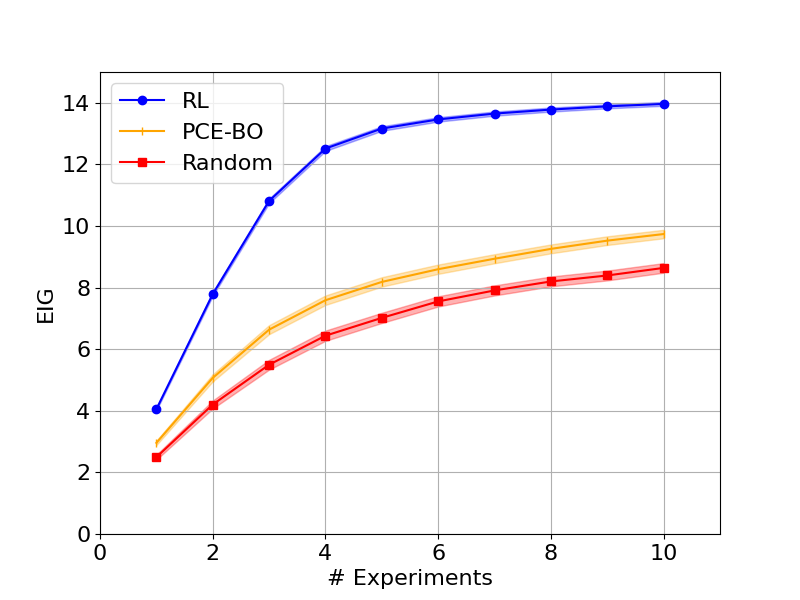}}
\caption{\gls{EIG} for the non-differentiable version of the \gls{CES} problem, estimated using \gls{SPCE} with $L=1e7$. Trendlines are means and shaded regions are standard errors aggregated from $2000$ rollouts (\gls{RL}) or $1000$ rollouts (PCE-BO and random).}
\label{fig:nondiff}
\end{center}
\vskip -0.2in
\end{figure}

The results of this experiment are shown in~\cref{fig:nondiff}. Note that the \gls{DAD} and \gls{VPCE} baselines are absent from the results, as both of these methods do not function in the absence of likelihood model gradients. Indeed, trying to run them with the \textit{no\textunderscore grad} statement results in an error as pytorch tries to perform automatic differentiation without a gradient. \gls{VPCE} has been replaced with a baseline that uses Bayesian optimization to maximize the myopic \gls{PCE} bound (PCE-BO). No such replacement has been made for \gls{DAD}, as this would involve high-dimensional \gls{BO} of the policy parameters, which is a challenging problem~\cite{greenhill2020bayesian}.

The performance of the \gls{RL} agents is nearly identical to what is seen in~\cref{fig:ces}. This is because we used the same random seeds, and because \gls{RL} does not rely on gradients from the likelihood model.  \gls{RL} outperforms all of the gradient-free baselines by a considerable margin, starting from the very first experiment, and is orders of magnitude faster at deployment time as per~\cref{table:time}.

\section{Related Work}

The classic approach to Bayesian optimal experimental design is to iteratively optimize the immediate \gls{EIG}, run the experiment, and infer a posterior. The \gls{EIG} has been estimated by particle filters~\cite{cavagnaro2010adaptive}, sequential Monte Carlo~\cite{drovandi2014sequential, moffat2020sequential}, nested Monte Carlo~\cite{myung2013tutorial}, multilevel Monte Carlo~\cite{goda2020multilevel}, Markov Chain Monte Carlo~\cite{muller2004optimal}, ratio estimation~\cite{kleinegesse2019efficient},  variational bounds~\cite{foster2019variational,foster2020unified}, neural estimation~\cite{kleinegesse2020bayesian}, Laplace importance sampling~\cite{beck2018fast} and more. Methods for specific models have been developed that exploit unique properties, such as in the case of linear models~\cite{verdinelli1996bayesian}, Gaussian Process models~\cite{pandita2021surrogate}, and polynomial models~\cite{rainforth2018nesting}.

Posterior inference has been done by sequential Monte Carlo~\cite{ moffat2020sequential, kleinegesse2021sequential}, Markov Chain Monte Carlo~\cite{myung2009optimal, huan2013simulation}, and variational inference~\cite{foster2020unified}.
More recently, alternatives have been proposed that avoid explicit computation of a posterior. These include adversarial methods~\cite{prangle2019bayesian, overstall2022properties}, and amortized methods~\cite{foster2021deep, ivanova2021implicit}.

Deep connections exist between the fields of experimental design and active learning.~\citet{gal2017deep} introduced the DBALD estimator for information gain, which is in fact a special case of \gls{PCE}~\cite{foster2022thesis}. Non-myopic methods have also been developed in the classification context, using n-step look-ahead~\cite{zhao2021uncertainty} or compressing the joint of multiple experiments into a one-dimensional integral~\cite{zhao2021efficient}.

The principles of optimal experimental design have also been applied in the context of reinforcement learning. ~\citet{sun2011planning} derived "curiosity Q-learning" based on the information gain quantity, but the resulting algorithm is only tractable in the case of tabular \gls{RL}, where the spaces of observations and actions are finite. ~\citet{houthooft2016vime} proposed to maximize the variational information of a predictive model as an auxiliary objective to the typical reward function. ~\citet{shyam2019model} explicitly use the information gain of an ensemble of world models to optimize an exploration policy in each iteration, which is then used to guide the next action in the real world. Finally, the field of active inference~\cite{friston2009reinforcement} seeks to solve \gls{RL} problems by minimizing the variational free energy, which is equivalent to maximizing the expected information gain of the posterior.

In contrast, this work seeks to go in the reverse direction, and apply \gls{RL} to solve experimental design problems. To the best of our knowledge, this is the first work that makes generic \gls{RL} algorithms applicable for optimal design of experiments. The work that comes nearest is~\citet{huan2016sequential,shen2021bayesian}, which proposes a specific Dynamic Programming algorithm and requires the inference of explicit posteriors, approximated on a discretized grid.

\section{Discussion}

In this work we proposed a method for transforming \acrfull{SED} problems into \acrfullpl{MDP} such that each pair of problems are solved by the same optimal policy. This allowed us to learn design policies using off-the-shelf \acrfull{RL} algorithms, resulting in an amortized algorithm in the spirit of \acrfull{DAD}. Our approach improves on the state-of-the-art in several important ways.

First, in the case where design spaces are continuous, our approach significantly outperformed all baselines, including both conventional methods that do not use amortization as well as the amortized \gls{DAD}. Our analyses indicate that this is due to the superior ability of \gls{RL} to explore the space of possible designs and thus discover better solutions. Indeed, investigation of the designs proposed by trained policies confirms that \gls{RL} policies have learned to propose much more diverse designs than the \gls{DAD} policies. However, the importance of exploration will vary between different \gls{SED} problems.
Second, in the case where design spaces are discrete, our method achieves comparable performance to the best non-amortized baselines. The amortized baselines simply do not work in this setting, because they depend on backpropagating gradients through the design.
Finally, in the case where the probabilistic model is not differentiable, the \gls{RL} approach achieves the best performance by far, whereas many of the baselines are inapplicable as they rely on gradient information which is not available. Thus our proposed method is applicable to a broad spectrum of design problems, is able to match or outperform the state-of-the-art in all of them, and still enjoys the fast deployment times of amortized methods that are restricted to continuous design spaces with differentiable probabilistic models.

Of note, our results cannot be achieved by naively using the \gls{SPCE} objective as a reward function. Ablations show that correct factorization of the rewards is critical to the performance of the resulting \gls{RL} policies.

Beyond the promising experimental results, it is important to note that our method opens the door to deploying \gls{RL} algorithms on sequential experimental design problems. The \gls{RL} literature is rich with techniques for handling problems with unique characteristics that make policy learning difficult, e.g. sparse rewards~\cite{zhang2021made} or high dimensionality~\cite{laskin2020curl}. Many of these techniques may prove to be useful for specific experimental design problems. With our approach, they can be applied in a straightforward manner by smartly transforming the design problem to an \gls{MDP}.

\bibliographystyle{icml2022}
\bibliography{main.bib}

\newpage
\appendix
\onecolumn
\setcounter{theorem}{0}

\section{Theorem Proofs}

In this appendix we provide proofs for the theorems in our paper. 

\setcounter{theorem}{0}

\subsection{Proof of~\cref{thm:naive_equivalence}}\label{proof:naive_equivalence}

Here we prove our first theorem, showing that even a naive formulation allows us to construct an equivalent \gls{hipmdp} to any sequential experiment optimization problem.

\begin{theorem}
    Let $\cM$ be a \gls{hipmdp} where $s_t = h_t; a_{t-1} = d_t \ \forall t \in [1,T]$ and initial state distribution and transition dynamics are $\rho_0 = h_0 \sim \delta(\emptyset); P_\Theta = \theta_{0:L} \sim p(\theta)$ and $\cT = p(y_t \g h_{t-1}, d_t, \theta_0)$ respectively. If $\gamma = 1$ and the reward function is:
    \begin{align}
        \cR(s_{t-1}, a_{t-1}, s_t, \theta) =
        \begin{cases}
            0,& \text{if } t < T \\
            g(\theta, h_t),& \text{if } t = T
        \end{cases}
    \end{align}
    
    then the expected return satisfies:
    \begin{equation}
        J(\pi) = sPCE(\pi, L, T).
    \end{equation}
\end{theorem}

\begin{proof}
    We need to show that $J(\pi) = sPCE(\pi, L, T)$ under the conditions of the theorem. Plugging in the definitions of return and \gls{SPCE}, this means showing that
    \begin{equation}
        \E_{\mathcal{T}, \pi, \rho_0, P_\Theta} \left[ \sum^T_{t=1} \gamma^{t-1} \cR(s_{t-1}, a_{t-1}, s_t, \theta) \right]
        = \E_{p(\theta_0, h_T \g \pi)p(\theta_{1:L})} \left[ \log \frac{p(h_T \g \theta_0, \pi)}{\frac{1}{L+1} \sum^L_{l=0} p(h_T \g \theta_l, \pi)} \right].
    \end{equation}
    
    Plugging in the definitions of $\cT, \pi, \rho_0, P_\Theta$, this is equivalent to
    \begin{align}
        &\E_{\prod_{t=1}^T \left[ p(y_t \g h_{t-1}, d_t, \theta_0) \pi(d_t \g h_{t-1}) \right] p(\theta_{0:L})} \left[ \sum^T_{t=1} \gamma^{t-1} \cR(s_{t-1}, a_{t-1}, s_t, \theta) \right]
        = \E_{p(\theta_0, h_T \g \pi)p(\theta_{1:L})} \left[ \log \frac{p(h_T \g \theta_0, \pi)}{\frac{1}{L+1} \sum^L_{l=0} p(h_T \g \theta_l, \pi)} \right] \\
        &\E_{p(h_T \g \theta_0, \pi) p(\theta_{0:L})} \left[ \sum^T_{t=1} \gamma^{t-1} \cR(s_{t-1}, a_{t-1}, s_t, \theta) \right]
        = \E_{p(\theta_0, h_T \g \pi)p(\theta_{1:L})} \left[ g(\theta, h_T) \right] \\
        &\E_{p(\theta_0, h_T \g \pi)p(\theta_{1:L})} \left[ \sum^T_{t=1} \gamma^{t-1} \cR(s_{t-1}, a_{t-1}, s_t, \theta) \right]
        = \E_{p(\theta_0, h_T \g \pi)p(\theta_{1:L})} \left[ g(\theta, h_T) \right].
    \end{align}
    Note that the distributions on both sides are identical. Therefore, if the term inside the expectation on the LHS is equal to the term inside the expectation on the RHS for all possible realizations of $h_T, \theta$ and $\pi$, then the expectations must be equal. Thus we seek to show that:
    \begin{equation}
        \sum^T_{t=1} \gamma^{t-1} \cR(s_{t-1}, a_{t-1}, s_t, \theta)
        = g(\theta, h_T).
    \end{equation}
    By the conditions of the theorem $\gamma = 1$ so this simplifies to:
    \begin{equation}
        \sum^T_{t=1} \cR(s_{t-1}, a_{t-1}, s_t, \theta)
        = g(\theta, h_T).
    \end{equation}
    Plugging in the definition of the reward function:
    \begin{equation}
        0 + 0 + \dots + g(\theta, h_T) = g(\theta, h_T),
    \end{equation}
    which is trivially true. Therefore $J(\pi) = sPCE(\pi, L, T)$.
\end{proof}

\subsection{Proof of~\cref{thm:equivalence}}\label{proof:equivalence}

In this appendix we prove the main theorem of the paper, which shows that we can construct an equivalent \gls{hipmdp} with dense rewards to any sequential experiment optimization problem. For convenience, we restate the theorem here along with some relevant definitions:
\begin{align}
    &B_{\psi,t} \equiv \sum_{k=1}^t ENC_\psi(d_k, y_k) \\
    &C_t \equiv \left[ \prod^t_{k=1} p(y_k \g \theta_l, d_k) \right]_{l=0}^L 
\end{align}

\begin{theorem}
    Let $\mathcal{M}$ be a \gls{hipmdp} where $s_t = (B_{\psi,t}, C_t, y_t); a_{t-1} = d_t \ \forall t \in [1,T]$ and initial state distribution is $\rho_0 = (\mathbf{0}, \mathbf{1}, \emptyset); P_\Theta = \theta_{0:L} \sim p(\theta)$, the reward function is
    \begin{align}
            \mathcal{R}(s_{t-1}, a_{t-1}, s_t, \theta) =& \log p(y_t \g \theta_0, d_t)
            - \log (C_t \cdot \mathbf{1}) + \log (C_{t-1} \cdot \mathbf{1}),
    \end{align}
    and transition function is
    \begin{align}
    \begin{split}
        y_t &\sim p(y_t \g d_t, \theta_0) \\
        B_{\psi,t} &= B_{\psi,t-1} + ENC_\psi(d_t, y_t) \\
        C_t &= C_{t-1} \odot \left[ p(y_t \g \theta_l, d_t) \right]_{l=0}^L.
    \end{split}
\end{align}
    If $\gamma = 1$ then the expected return satisfies:
    \begin{equation}
        J(\pi) = sPCE(\pi, L, T).
    \end{equation}
\end{theorem}

\begin{proof}
    We need to show that $J(\pi) = sPCE(\pi, L, T)$ under the conditions of the theorem. Plugging in the definitions of return and \gls{SPCE}, this means showing that
    \begin{equation}
        \E_{\mathcal{T}, \pi, \rho_0, P_\Theta} \left[ \sum^T_{t=1} \gamma^{t-1} \cR(s_{t-1}, a_{t-1}, s_t, \theta) \right]
        = \E_{p(\theta_0, h_T \g \pi)p(\theta_{1:L})} \left[ \log \frac{p(h_T \g \theta_0, \pi)}{\frac{1}{L+1} \sum^L_{l=0} p(h_T \g \theta_l, \pi)} \right].
    \end{equation}
    Plugging in the definitions of $\mathcal{T}, \pi, \rho_0, P_\Theta$ we get:
    \begin{align}
        \begin{split}
            &\E_{\prod_{t=1}^T \left[ p(B_{\psi,t}, C_t, y_t \g B_{\psi,t-1}, C_{t-1}, y_{t-1}, d_t, \theta_0) \pi(d_t \g B_{\psi,t-1}) \right] p(\theta_{0:L})} \left[ \sum^T_{t=1} \gamma^{t-1} \cR(s_{t-1}, a_{t-1}, s_t, \theta) \right] \\
            &= \E_{p(\theta_0, h_T \g \pi)p(\theta_{1:L})} \left[ \log \frac{p(h_T \g \theta_0, \pi)}{\frac{1}{L+1} \sum^L_{l=0} p(h_T \g \theta_l, \pi)} \right]
        \end{split} \\
        \begin{split}
            &\E_{\prod_{t=1}^T \left[ p(B_{\psi,t}, C_t \g B_{\psi,t-1}, C_{t-1}, d_t, y_t) p(y_t \g d_t, \theta_0) \pi(d_t \g B_{\psi,t-1}) \right] p(\theta_{0:L})} \left[ \sum^T_{t=1} \gamma^{t-1} \cR(s_{t-1}, a_{t-1}, s_t, \theta) \right] \\
            &= \E_{p(\theta_0, h_T \g \pi)p(\theta_{1:L})} \left[ \log \frac{p(h_T \g \theta_0, \pi)}{\frac{1}{L+1} \sum^L_{l=0} p(h_T \g \theta_l, \pi)} \right]
        \end{split}.
    \end{align}
    Noting that $B_{\psi,t}$ and $C_t$ can be computed exactly from $h_t$, we can simplify the LHS further:
    \begin{align}
        \begin{split}
            &\E_{\prod_{t=1}^T \left[ p(h_t \g h_{t-1}, d_t, y_t) p(y_t \g d_t, \theta_0) \pi(d_t \g h_{t-1}) \right] p(\theta_{0:L})} \left[ \sum^T_{t=1} \gamma^{t-1} \cR(s_{t-1}, a_{t-1}, s_t, \theta) \right] \\
            & = \E_{p(\theta_0, h_T \g \pi)p(\theta_{1:L})} \left[ \log \frac{p(h_T \g \theta_0, \pi)}{\frac{1}{L+1} \sum^L_{l=0} p(h_T \g \theta_l, \pi)} \right]
        \end{split}
         \\
        &\E_{p(h_T \g \theta_0, \pi) p (\theta_{0:L})} \left[ \sum^T_{t=1} \gamma^{t-1} \cR(s_{t-1}, a_{t-1}, s_t, \theta) \right] 
        = \E_{p(\theta_0, h_T \g \pi)p(\theta_{1:L})} \left[ \log \frac{p(h_T \g \theta_0, \pi)}{\frac{1}{L+1} \sum^L_{l=0} p(h_T \g \theta_l, \pi)} \right] \\
        &\E_{p(\theta_0, h_T \g \pi)p(\theta_{1:L})} \left[ \sum^T_{t=1} \gamma^{t-1} \cR(s_{t-1}, a_{t-1}, s_t, \theta) \right] 
        = \E_{p(\theta_0, h_T \g \pi)p(\theta_{1:L})} \left[ \log \frac{p(h_T \g \theta_0, \pi)}{\frac{1}{L+1} \sum^L_{l=0} p(h_T \g \theta_l, \pi)} \right].
    \end{align}
    Note that the distributions on both sides are identical. Therefore, if the term inside the expectation on the LHS is equal to the term inside the expectation on the RHS for all possible realizations of $h_T, \theta$ and $\pi$, then the expectations must be equal. Thus we seek to show that:
    \begin{equation}
           \sum^T_{t=1} \gamma^{t-1} \cR(s_{t-1}, a_{t-1}, s_t, \theta) = \log \frac{p(h_T \g \theta_0, \pi)}{\frac{1}{L+1} \sum^L_{l=0} p(h_T \g \theta_l, \pi)}. \label{eq:inner_equivalence}
    \end{equation}
    Recall that we have made the same assumption as~\citet{foster2021deep} that $p(h_T \g \theta, \pi) = \prod^T_{t=1} p(h_t \g \theta_0, \pi, h_{t-1})$. We can therefore rewrite the RHS:
    \begin{align}
         RHS &= \log \frac{p(h_T \g \theta_0, \pi)}{\frac{1}{L+1} \sum^L_{l=0} p(h_T \g \theta_l, \pi)} \\
        &= \log \frac{\prod^T_{t=1} p(h_t \g \theta_0, \pi, h_{t-1})}{\frac{1}{L+1} \sum^L_{l=0} \prod^T_{t=1} p(h_t \g \theta_l, \pi, h_{t-1})} \\
        &= \log \frac{\prod^T_{t=1} p(y_t \g \theta_0, d_t) \pi(d_t \g h_{t-1})}{\frac{1}{L+1} \sum^L_{l=0} \prod^T_{t=1} p(y_t \g \theta_l, d_t) \pi(d_t \g h_{t-1})} \\
        &= \log \frac{\prod^T_{t=1} p(y_t \g \theta_0, d_t)}{\frac{1}{L+1} \sum^L_{l=0} \prod^T_{t=1} p(y_t \g \theta_l, d_t)},
    \end{align}
    where the last line uses the fact that $\pi(d_t \g h_{t-1})$ is independent of $\theta$. Applying log rules, we get:
    \begin{align}
        &= \sum^T_{t=1} \log p(y_t \g \theta_0, d_t) - \log \frac{1}{L+1} \sum^L_{l=0} \prod^T_{t=1} p(y_t \g \theta_l, d_t) \\
        &= \sum^T_{t=1} \log p(y_t \g \theta_0, d_t) - \log \frac{1}{L+1} - \log \sum^L_{l=0} \prod^T_{t=1} p(y_t \g \theta_l, d_t) \\
        \begin{split}
            &= \sum^T_{t=1} \log p(y_t \g \theta_0, d_t) + \log (L+1) - \sum^T_{t=1} \left[ \log \sum^L_{l=0} \prod^t_{k=1} p(y_k \g \theta_l, d_k) - \log \sum^L_{l=0} \prod^{t-1}_{k=1} p(y_k \g \theta_l, d_k) \right] \\
            &\quad - \log \sum^L_{l=0} \prod^0_{k=1} p(y_k \g \theta_l, d_k),
        \end{split}
    \end{align}
    where the last step breaks $\log \sum^L_{l=0} \prod^T_{t=1} p(y_t \g \theta_l, d_t)$ into a telescoping sum where the terms cancel out. Now, noting that the product of an empty sequence is by definition $1$ we get:
    \begin{align}
        &= \sum^T_{t=1} \log p(y_t \g \theta_0, d_t) + \log (L+1) - \sum^T_{t=1} \left[ \log \sum^L_{l=0} \prod^t_{k=1} p(y_k \g \theta_l, d_k) - \log \sum^L_{l=0} \prod^{t-1}_{k=1} p(y_k \g \theta_l, d_k) \right] - \log (L+1) \\
        &= \sum^T_{t=1} \log p(y_t \g \theta_0, d_t) - \sum^T_{t=1} \left[ \log \sum^L_{l=0} \prod^t_{k=1} p(y_k \g \theta_l, d_k) - \log \sum^L_{l=0} \prod^{t-1}_{k=1} p(y_k \g \theta_l, d_k) \right] \\
        &= \sum^T_{t=1} \left[ \log p(y_t \g \theta_0, d_t) - \log \sum^L_{l=0} \prod^t_{k=1} p(y_k \g \theta_l, d_k) + \log \sum^L_{l=0} \prod^{t-1}_{k=1} p(y_k \g \theta_l, d_k) \right]. \label{eq:reward}
    \end{align}
    
    By the conditions of the theorem, we have that
    \begin{align}
        &\gamma = 1 \\
        &\mathcal{R}(s_{t-1}, a_{t-1}, s_t, \theta) = \log p(y_t \g \theta_0, d_t) 
            - \log \sum^L_{l=0} \prod^t_{k=1} p(y_k \g \theta_l, d_k) + \log \sum^L_{l=0} \prod^{t-1}_{k=1} p(y_k \g \theta_l, d_k).
    \end{align}
    
    Plugging this into~\cref{eq:reward}, we get:
    \begin{align}
        RHS &= \sum^T_{t=1} \mathcal{R}(s_{t-1}, a_{t-1}, s_t, \theta) \\
        RHS &= \sum^T_{t=1} 1^{t-1} \mathcal{R}(s_{t-1}, a_{t-1}, s_t, \theta) \\
        \log \frac{p(h_T \g \theta_0, \pi)}{\frac{1}{L+1} \sum^L_{l=0} p(h_T \g \theta_l, \pi)} &= \sum^T_{t=1} \gamma^{t-1} \mathcal{R}(s_{t-1}, a_{t-1}, s_t, \theta).
    \end{align}
    Which is the equivalence we sought to show in~\cref{eq:inner_equivalence}. Therefore $J(\pi) = sPCE(\pi, L, T)$.
\end{proof}

\subsection{Q-Learning from Priors} \label{proof:q_learning}

In this appendix we prove an additional theorem that shows we can learn Q-function approximators without needing to infer posteriors. It is based in part on a related proof from Section 2.1.3 of~\citet{guez2015sample}, which establishes a recursion for the state value function in Bayes-Adaptive MDPs. However,~\citet{guez2015sample} does not show that the value function can be computed using only samples from the prior.

\begin{theorem}
    Let $\cM$ be a \gls{hipmdp} where $s_t = f_1(h_t);\ a_t = f_2(d_{t+1})$ for some deterministic functions $f_1, f_2$ and the transition dynamics are some deterministic function of $y_t \sim p(y_t \g h_{t-1}, d_t, \theta)$. For an arbitrary policy $\pi$, the Q-function estimator $\hat{Q}_\pi$ can be learned using trajectories sampled from the joint $p(\theta, h_T \g \pi)$
\end{theorem}

\begin{proof}
    To begin the proof we must first derive a Bellman equation for the Q-function in the \gls{hipmdp} $\cM$. In other words, we need to show that
    
    \begin{equation}
        Q_\pi(s_i, a_i) = \E \left[ \cR(s_i, a_i, s_{i+1}, \theta) + Q_\pi(s_{i+1}, a_{i+1})  \right]
    \end{equation}
    for some expectation. Since the state and action are deterministic functions of the history and design, respectively, we can without loss of generality replace $s_t$ with $h_t$ and $a_t$ with $d_{t+1}$ in the following derivation, with the functions $f_1$ and $f_2$ being implicitly subsumed into the reward and value functions. Thus we seek to show that
    
    \begin{equation}
        Q_\pi(h_i, d_{i+1}) = \E \left[ \cR(h_{i+1}, \theta) + Q_\pi(h_{i+1}, d_{i+2})  \right]. \label{eq:generic_recursion}
    \end{equation}
    
    Recall that a \gls{hipmdp} is a distribution over \glspl{MDP}, where a particular realization of $\theta$ defines an \gls{MDP} sampled from that distribution. We therefore define the state-action-parameter value function $U_\pi$:
    
    \begin{align}
    U_\pi(h_i, d_{i+1}, \theta) =& \E_{\prod_{t=i+1}^T \left[ y_t \sim p(y_t \g d_t, \theta); d_{t+1} \sim \pi(d_{t+1} \g h_t) \right]}
        \left[ \sum_{t=i}^T \gamma^{t-i} \cR(d_{1:t+1}, y_{1:t+1}, \theta) \right] \\
        \begin{split}
        =& \E_{y_{i+1} \sim p(y_{i+1} \g d_{i+1}, \theta); d_{i+2} \sim \pi(d_{i+2} \g h_{i+1})} \Bigg[ \cR(d_{1:i+1}, y_{1:i+1}, \theta) \\
        &+ \gamma \cdot \E_{\prod_{t=i+2}^T \left[  y_t \sim p(y_t \g d_t, \theta); d_{t+1} \sim \pi(d_{t+1} \g h_t) \right]} \left[ \sum_{t=i+1}^T \gamma^{t-i} \cR(d_{1:t+1}, y_{1:t+1}, \theta) \right] \Bigg]
        \end{split}\\
        =& \E_{y_{i+1} \sim p(y_{i+1} \g d_{i+1}, \theta); d_{i+2} \sim \pi(d_{i+2} \g h_{i+1})} \left[ \cR(d_{1:i+1}, y_{1:i+1}, \theta) + \gamma \cdot U_\pi(h_{i+1}, d_{i+2}, \theta) \right],
    \end{align}
    and we note that this is equivalent to the state-action value function in the \gls{MDP} defined by $\theta$. By construction, the Q-function is the expectation of the U-function:
    
    \begin{equation}
        Q_\pi(h_i, d_{i+1}) = \E_{\theta \sim p(\theta \g h_i)} U_\pi(h_i, d_{i+1}, \theta).
    \end{equation}
    
    Plugging in the recursive relationship of $U_\pi$ into this expression, we get
    
    \begin{align}
    Q_\pi(h_i, d_{i+1}) =& \E_{\theta \sim p(\theta \g h_i); y_{i+1} \sim p(y_{i+1} \g d_{i+1}, \theta); d_{i+2} \sim \pi(d_{i+2} \g h_{i+1})} \left[ \cR(d_{1:i+1}, y_{1:i+1}, \theta) + \gamma \cdot U_\pi(h_{i+1}, d_{i+2}, \theta) \right] \\
    \begin{split}
        =& \E_{\theta \sim p(\theta \g h_i); y_{i+1} \sim p(y_{i+1} \g d_{i+1}, \theta); d_{i+2} \sim \pi(d_{i+2} \g h_{i+1})} \left[ \cR(d_{1:i+1}, y_{1:i+1}, \theta) \right] \\
        &+ \gamma \cdot \E_{\theta \sim p(\theta \g h_i); y_{i+1} \sim p(y_{i+1} \g d_{i+1}, \theta); d_{i+2} \sim \pi(d_{i+2} \g h_{i+1})} [ U_\pi(h_{i+1}, d_{i+2}, \theta) ]. \label{eq:first_recursion}
    \end{split}
    \end{align}
    
    We want a recursive expression for the Q-function, so we need to show that $U_\pi(h_{i+1}, d_{i+2}, \theta)$ is equal in expectation to $Q_\pi(h_{i+1}, d_{i+2})$. Using the identity $p(\theta \g h_i) p(y_{i+1} \g d_{i+1}, \theta) = p(\theta \g h_{i+1}) p(y_{i+1} \g h_i, d_{i+1})$ we get:
    
    \begin{align}
        \E_{p(\theta \g h_i) p(y_{i+1} \g d_{i+1}, \theta) \pi(d_{i+2} \g h_{i+1}) } [ U_\pi(h_{i+1}, d_{i+2}, \theta) ]
        &= \E_{p(\theta \g h_{i+1}) p(y_{i+1} \g h_i, d_{i+1}) \pi(d_{i+2} \g h_{i+1})}  [ U_\pi(h_{i+1}, d_{i+2}, \theta) ] \\
        &= \E_{p(y_{i+1} \g h_i, d_{i+1}) \pi(d_{i+2} \g h_{i+1})} \E_{p(\theta \g h_{i+1})}  [ U_\pi(h_{i+1}, d_{i+2}, \theta) ] \\
        &= \E_{p(y_{i+1} \g h_i, d_{i+1}) \pi(d_{i+2} \g h_{i+1})} [ Q_\pi(h_{i+1}, d_{i+2}) ] \\
        &= \E_{p(y_{i+1} \g h_i, d_{i+1}) \pi(d_{i+2} \g h_{i+1})} \E_{p(\theta \g h_{i+1})} [ Q_\pi(h_{i+1}, d_{i+2}) ] \\
        &= \E_{p(\theta \g h_i) p(y_{i+1} \g d_{i+1}, \theta) \pi(d_{i+2} \g h_{i+1})} \left[ Q_\pi(h_{i+1}, d_{i+2}) \right].
    \end{align}

    Now we plug this back into~\cref{eq:first_recursion}:
    
    \begin{align}
        \begin{split}
            Q_\pi(h_i, d_{i+1}) = &\E_{p(\theta \g h_i) p(y_{i+1} \g d_{i+1}, \theta) \pi(d_{i+2} \g h_{i+1}) } \left[ \cR(d_{1:i+1}, y_{1:i+1}, \theta) \right] \\
            &+ \gamma \cdot \E_{p(\theta \g h_i) p(y_{i+1} \g d_{i+1}, \theta) \pi(d_{i+2} \g h_{i+1}) } \left[ U_\pi(h_{i+1}, d_{i+2}, \theta) \right]
        \end{split} \\
        \begin{split}
            = &\E_{p(\theta \g h_i) p(y_{i+1} \g d_{i+1}, \theta) \pi(d_{i+2} \g h_{i+1})} \left[ \cR(d_{1:i+1}, y_{1:i+1}, \theta) \right] \\
            &+ \gamma \cdot \E_{p(\theta \g h_i) p(y_{i+1} \g d_{i+1}, \theta) \pi(d_{i+2} \g h_{i+1})} \left[ Q_\pi(h_{i+1}, d_{i+2}) \right]
        \end{split} \\
        = &\E_{p(\theta \g h_i) p(y_{i+1} \g d_{i+1}, \theta) \pi(d_{i+2} \g h_{i+1})} \left[ \cR(d_{1:i+1}, y_{1:i+1}, \theta) + Q_\pi(h_{i+1}, d_{i+1}) \right],
    \end{align}
    which gives us a recursive expression of the form in~\cref{eq:generic_recursion}. To train the estimator $\hat{Q}_\pi$ we need to minimize some expected loss w.r.t. the distribution of trajectories generated by the policy, i.e.
    
    \begin{equation}
        \E_{p(h_i, d_{i+1} \g \pi)} f_{loss}(\hat{Q}_\pi(h_i, d_{i+1}), Q_\pi(h_i, d_{i+1})),
    \end{equation}
    where the loss is a deterministic function, for example the mean square error. We can use Monte Carlo sampling to estimate this expectation. $\hat{Q}_\pi$ is available in functional form (e.g. as a neural net) and can be easily evaluated for any input. To evaluate $Q_\pi$, however, we would have to sample from the distribution $p(h_i, d_{i+1} \g \pi) p(\theta \g h_i) p(y_{i+1} \g d_{i+1}, \theta)  \pi(d_{i+2} \g h_{i+1})$. This requires sampling from several difficult distributions, such as the marginal $p(h_i, d_i \g \pi)$ and the posterior $p(\theta \g h_i)$. However, recalling the identity $p(\theta \g h_i) p(y_{i+1} \g d_{i+1}, \theta) = p(\theta \g h_{i+1}) p(y_{i+1} \g h_i, d_{i+1})$, we can simplify the above distribution:
    
    \begin{align}
       p(h_i, d_{i+1} \g \pi) p(\theta \g h_i) p(y_{i+1} \g d_{i+1}, \theta)  \pi(d_{i+2} & \g h_{i+1}) = \\
            &= p(\theta \g h_i) p(y_{i+1} \g d_{i+1}, \theta) p(h_i, d_{i+1} \g \pi) \pi(d_{i+2} \g h_{i+1}) \\
            &= p(\theta \g h_i) p(y_{i+1} \g d_{i+1}, \theta) p(d_{i+1} \g h_i, \pi) p(h_i \g \pi) \pi(d_{i+2} \g h_{i+1}) \\
            &=  p(\theta \g h_{i+1}) p(y_{i+1} \g h_i, d_{i+1}) p(d_{i+1} \g h_i, \pi) p(h_i \g \pi) \pi(d_{i+2} \g h_{i+1}) \\
            &=  p(\theta \g h_{i+1}) p(y_{i+1}, d_{i+1} \g h_i, \pi) p(h_i \g \pi) \pi(d_{i+2} \g h_{i+1}) \\
            &=  p(\theta \g h_{i+1}) p(h_{i+1} \g h_i, \pi) p(h_i \g \pi) \pi(d_{i+2} \g h_{i+1}) \\
            &= p(\theta \g h_{i+1}) p(h_{i+1}, h_i \g \pi) \pi(d_{i+2} \g h_{i+1}) \\
            &= p(\theta \g h_{i+1}) p(h_{i+1}\g \pi) \pi(d_{i+2} \g h_{i+1}) \\
            &= p(\theta, h_{i+1} \g \pi) \pi(d_{i+2} \g h_{i+1}).
    \end{align}
    
    Hence we can generate the necessary samples by drawing $\theta$ from the prior and rolling out the policy, i.e. sampling from the joint $p(\theta, h_T \g \pi)$.
    
\end{proof}

\section{Experiment Details}

In this section we include the full description of all the experimental design problems considered in the paper.

\subsection{Source Location}\label{app:source}

In this experiment there are $2$ sources embedded in the $2$-dimensional plane that emit a signal. The total intensity at any given coordinate $d$ in the plane is:
\begin{equation}
    \mu(\theta, d) = b + \frac{1}{m + ||\theta_1 - d||^2}  + \frac{1}{m + ||\theta_2 - d||^2},
\end{equation}
where $b, m > 0$ are the background and maximum signals, respectively, $||\cdot||^2$ is the squared Euclidean norm, and $\theta_i$ are the coordinates of the $i^{th}$ signal source. The probabilistic model is:
\begin{equation}
    \theta_i \sim \cN(0, I); \quad \log y \g \theta, d \sim \cN(log(\mu(\theta, d), \sigma),
\end{equation}
i.e. the prior is unit Gaussian and we observe the log of the total signal intensity with some Gaussian observation noise $\sigma$.
The design space is restricted to $\cA = [-4, 4]^2$. The hyperparameters we used are
\begin{table}[h]
\vskip 0.15in
\begin{center}
\begin{small}
\begin{sc}
\begin{tabular}{lr}
\toprule
Parameter & Value \\
\midrule
$b$    & $1e-1$  \\
$m$    & $1e-4$  \\
$\sigma$    & $0.5$  \\
\bottomrule
\end{tabular}
\end{sc}
\end{small}
\end{center}
\vskip -0.1in
\end{table}

\subsection{Constant Elasticity of Substitution}\label{app:ces}

In this experiment subjects compare $2$ baskets of goods and give a rating on a sliding scale from $0$ to $1$. The designs are vectors $d = (x, x^\prime)$ where $x, x^\prime \in [0, 100]^3$ are the baskets of goods. The latent parameters and priors are:

\begin{align}
    \rho &\sim Beta(1,1)\\
    \alpha &\sim Dirichlet([1,1,1])\\
    \log u &\sim \cN(1,3).
\end{align}

The probabilistic model is:

\begin{align}
    U(x) &= \left(\sum_i x^\rho_i\alpha_i\right)^{1/\rho}\\
    \mu_\eta &= (U(x) - U(x^\prime)) u\\
    \sigma_\eta &= (1 + ||x - x^\prime||) \tau\cdot u\\
    \eta &\sim \cN(\mu_\eta, \sigma^2_\eta)\\
    y &= clip(sigmoid(\eta), \epsilon, 1-\epsilon),
\end{align}
where $U(\cdot)$ is a subjective utility function. The hyperparameter values are:
\begin{table}[h]
\vskip 0.15in
\begin{center}
\begin{small}
\begin{sc}
\begin{tabular}{lr}
\toprule
Parameter & Value \\
\midrule
$\tau$    & $0.005$  \\
$\epsilon$    & $2^{-22}$  \\
\bottomrule
\end{tabular}
\end{sc}
\end{small}
\end{center}
\vskip -0.1in
\end{table}

\subsection{Prey Population}\label{app:prey}

In this experiment an initial population of prey animals is left to survive for $\cT=24$ hours, and we measure the number of individuals consumed by predators at the end of the experiment. The designs are the initial populations $d = N_0 \in [1,300]$. The latent parameters and priors are:
\begin{align}
    \log a &\sim \cN(-1.4, 1.35)\\
    \log T_h &\sim \cN(-1.4, 1.35).
\end{align}

The population changes over time according to the differential equation:
\begin{equation}
    \frac{dN}{d\tau} = - \frac{aN^2}{1 + aT_hN^2}. 
\end{equation}

And the population $N_\cT$ is thus the solution of an initial value problem. The probabilistic model is:
\begin{align}
    p_\cT &= \frac{d - N_\cT}{d}\\
    y &\sim Binom(d, p_\cT).
\end{align}

\section{Algorithm Details}\label{app:details}

In this section we provide implementation details and hyperparameter settings for the various algorithms used in the paper. All of our code was written in Python using the Pytorch framework for automatic differentiation.

\subsection{RL}

The \gls{REDQ}~\cite{chen2020randomized} algorithm is a generalisation of the earlier Soft Actor-Critic, where instead of $2$ critic networks we train an ensemble of $N$ critics and estimate the Q value by a random sample of $M < N$ members from this ensemble. We therefore modified the implementation of Soft Actor-Critic available in Garage (\url{https://github.com/rlworkgroup/garage}). Hyparparameter choices for the different \gls{SED} problems are enumerated in the table below. Values were chosen by linear search over each hyperparameter independently, with the following search spaces:
target update rate $\tau \in \{1e-3, 5e-3, 1e-2\}$; policy learning rate $LR_\pi \in \{1e-5, 1e-4, 3e-4, 1e-3\}$; q-function learning rate $LR_{qf} \in \{1e-5, 1e-4, 3e-4, 1e-3\}$; replay buffer size $|buffer| \in \{1e5, 1e6, 1e7\}$; discount factor $\gamma \in \{0.95, 0.99, 1\}$. Note that we chose $\gamma \neq 1$ as we empirically observed this leads to improved performance.

\begin{table}[h]
\vskip 0.15in
\begin{center}
\begin{small}
\begin{sc}
\begin{tabular}{llll}
\toprule
Parameter & Source Location & CES & Prey Population \\
\midrule
$N$    & $2$ & $2$ &  $10$ \\
$M$    & $2$ & $2$ & $2$  \\
training iterations    & $2e4$ & $2e4$  & $4e4$  \\
contrastive samples    & $1e5$ & $1e5$ & $1e4$  \\
$T$    & $30$ & $10$ & $10$  \\
$\gamma$    & $0.9$ & $0.9$  & $0.95$  \\
$\tau$    & $1e-3$ & $5e-3$ & $1e-2$  \\
policy learning rate    & $1e-4$ & $3e-4$ & $1e-4$  \\
critic learning rate    & $3e-4$ & $3e-4$ & $1e-3$  \\
buffer size    & $1e7$ & $1e6$ & $1e6$  \\
\bottomrule
\end{tabular}
\end{sc}
\end{small}
\end{center}
\vskip -0.1in
\end{table}

The summary network $B_\psi$ consisted of $2$ fully connected layers with $128$ units and ReLU activation, followed by an output layer with $64$ units and no activation function. The policy network $\pi_\phi$ has a similar architecture, but its output layer depends on the experiment.
For the source location and CES experiments, the output layer has $2 |\cA|$ dimensions, representing the mean and log-variance of $|\cA|$ independent Tanh-Gaussian distributions, one for each dimension of the design space. For the prey population experiment, the output layer has $|\cA|$ dimensions, representing the logits of a Gumbel-Softmax distribution with $|\cA|$ classes.

\subsection{DAD}

We used the official \gls{DAD} implementation published by~\citet{foster2021deep} and available at \url{https://github.com/ae-foster/dad}. We verified that the authors' evaluation code produces the same results as ours for the same trained \gls{DAD} policy, and subsequently used their code to evaluate all \gls{DAD} policies.

\subsection{SMC}

For \gls{SMC} we used the implementation of~\citet{moffat2020sequential} (\url{https://github.com/haydenmoffat/sequential_design_for_predator_prey_experiments}). We connected to the R code by means of the $rpy2$ Python package. All hyperparameters are identical to the original paper. We used the parameter estimation utility and the model was Holling's type-II Binomial.

\subsection{VPCE}

\gls{VPCE} consists of maximizing the \gls{PCE} lower bound by gradient descent w.r.t.~the experimental design and inferring the posterior through variational inference, then repeating the process. We ran this algorithm with the hyperparameters in the table below. Hyperparameters were chosen based on the recommendations of~\citet{foster2020unified} where relevant.

\begin{table}[h]
\vskip 0.15in
\begin{center}
\begin{small}
\begin{sc}
\begin{tabular}{llll}
\toprule
Parameter & Source Location & CES & Prey Population \\
\midrule
$T$    & $30$ & $10$ & $10$  \\
design gradient steps    & $2500$ & $2500$ &  $2500$ \\
contrastive samples    & $500$ & $10$ & $100$  \\
expectation samples    & $500$ & $10$ & $100$  \\
VI samples    & $10$ & $10$  & $10$  \\
VI gradient steps   & $1000$ & $1000$ & $1000$  \\
VI learning rate    & $4e-2$ & $4e-2$ & $4e-2$  \\
\bottomrule
\end{tabular}
\end{sc}
\end{small}
\end{center}
\vskip -0.1in
\end{table}

For PCE-BO, we used a Matern $\frac{5}{2}$ kernel with a lengthscale of $10$ and $4$ steps of \gls{BO}.

\section{Hardware details}

Except for \gls{SMC}, all experiments were run on a Slurm cluster node with a single Nvidia Tesla P100 GPU and $4$ cores of an Intel Xeon E5-2690 CPU. The SMC experiments did not run on a GPU as the original authors' R code is not written to utilize GPU acceleration.

\section{Non-Myopic Solutions}

In general it is not feasible to determine the myopically optimal solution to a \gls{SED} problem, and therefore it is unclear whether an agent is actually finding a non-myopic solution or just a better myopic solution than the baselines.~\citet{foster2021deep} proposed to use a simplified variant of the source location problem, i.e. setting $d=1; k=1; T=2$. In this variant the \gls{EIG} of a design can be computed through numerical integration, and an optimal design can be approximated with high accuracy via grid search. We executed an experiment in this setting with $2$ \gls{RL} agents: a myopic agent (derived by setting $\gamma = 0$) and a non-myopic agent ($\gamma = 1$).

\begin{figure}[h] 
    \centerline{\includegraphics[width=0.95\columnwidth]{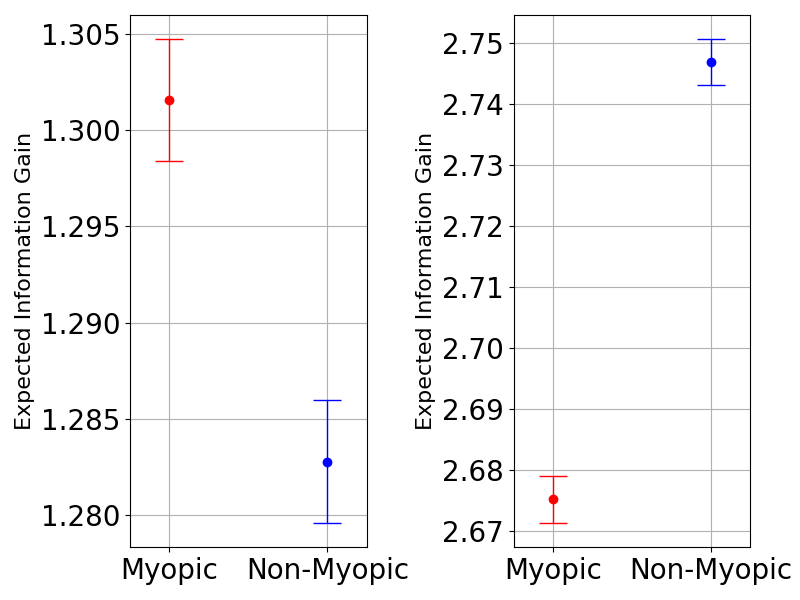}}
\vskip -0.2in
\caption{Mean and standard error of expected information gain for the $1$-dimensional source location problem, aggregated from $1e5$ rollouts with $L=1e4$ contrastive samples. (Left) information gain after the first experiment. (Right) information gain after the second experiment.} \label{fig:toy_experiment}
\end{figure}

The results are shown in~\cref{fig:toy_experiment}. Note that after the second experiment, our myopic agent achieves a very similar result to the optimal myopic strategy of~\citet{foster2021deep}, and the non-myopic agent outperforms both by a significant margin. However, at $t=1$ the non-myopic agent does worse than the myopic agent. This shows that the agent has indeed learned to sacrifice \gls{EIG} in earlier experiments in order to set itself up for superior \gls{EIG} at $t=T$.

\section{Learning Curves}

Plotted below are the learning curves for the reinforcement learning agents in the $3$ \gls{SED} problems investigated in this paper. The plots include mean and standard errors of the rewards for both the full method and the naive baseline, aggregated over $10$ replications. Note that the reward is computed with a smaller number of contrastive samples than was used in the final evaluation. For exact values cf.~\cref{app:details}.

These learning curves illustrate how our factorized reward formulation provides a more informative learning signal to the agent, allowing it to improve its performance faster (in a smaller number of iterations) and converge to higher rewards. This is particularly notable in the source location problem, where the high experimental budget ($30$ experiments as opposed to only $10$ in the other tasks) makes the problem of credit assignment exceptionally challenging.

\begin{figure}[h]
    \centerline{\includegraphics[width=0.8\columnwidth]{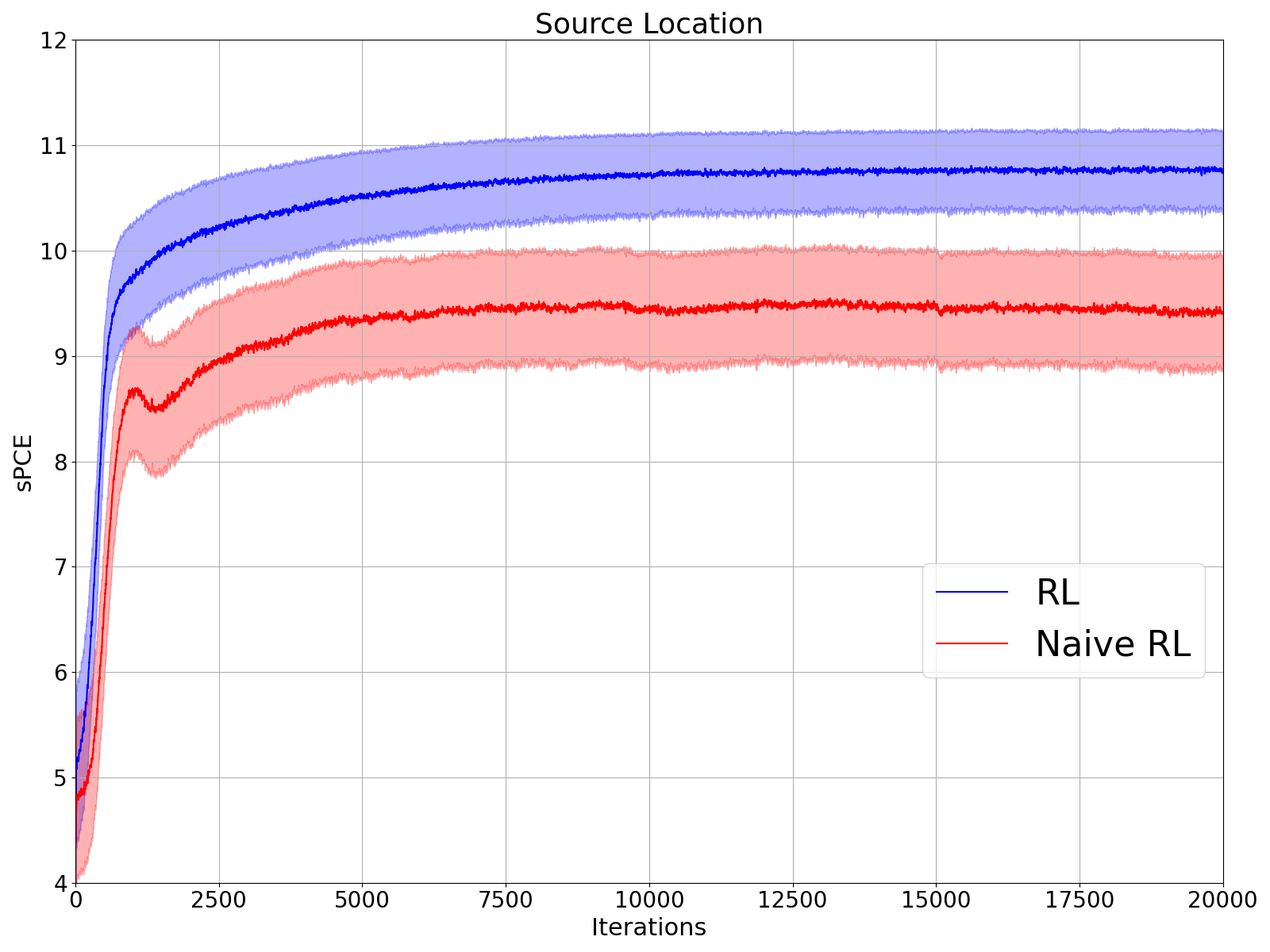}}
\vskip -0.2in
\end{figure}
\begin{figure}[h]
    \centerline{\includegraphics[width=0.8\columnwidth]{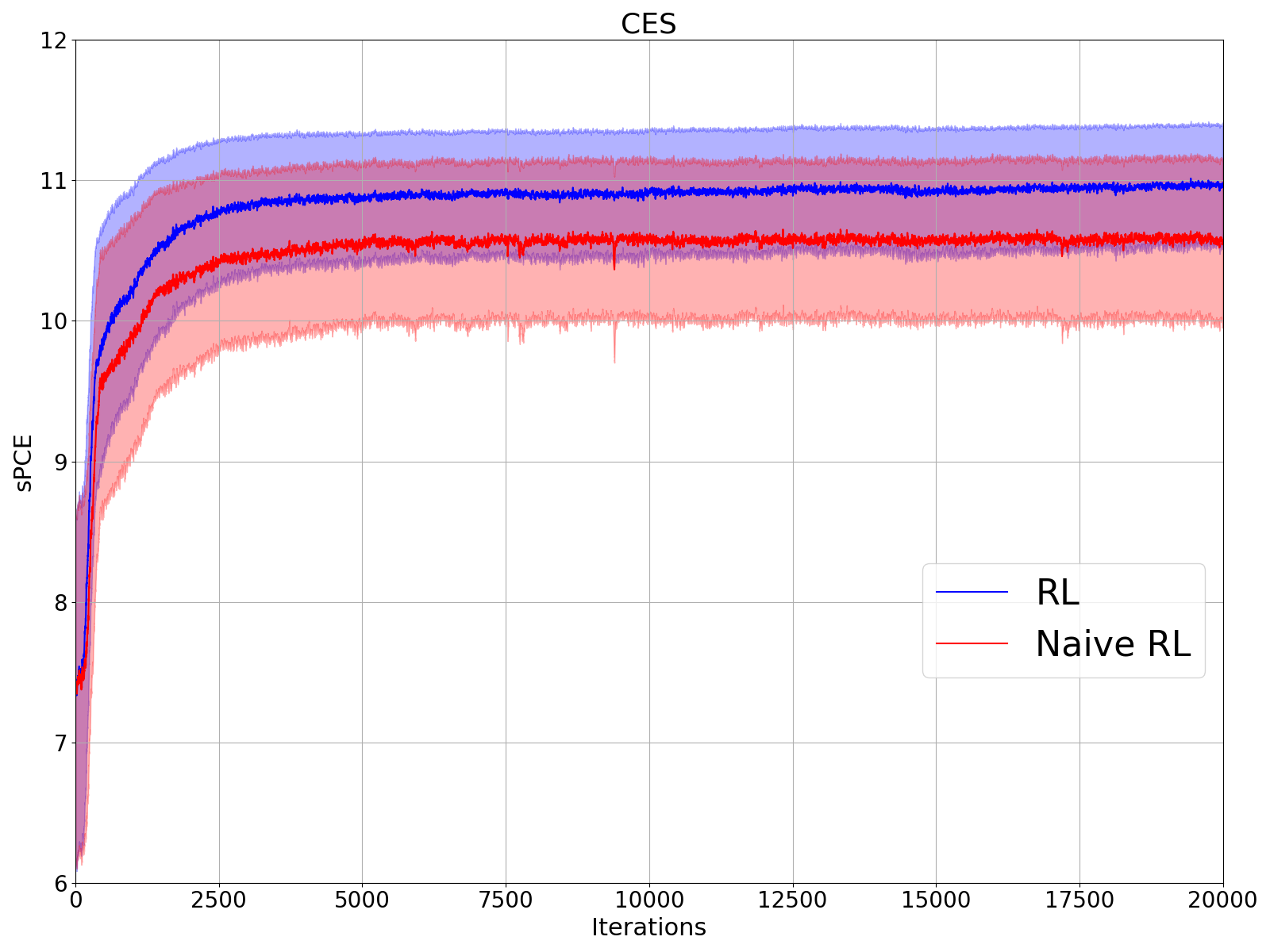}}
\vskip -0.2in
\end{figure}
\begin{figure}[h]
    \centerline{\includegraphics[width=0.8\columnwidth]{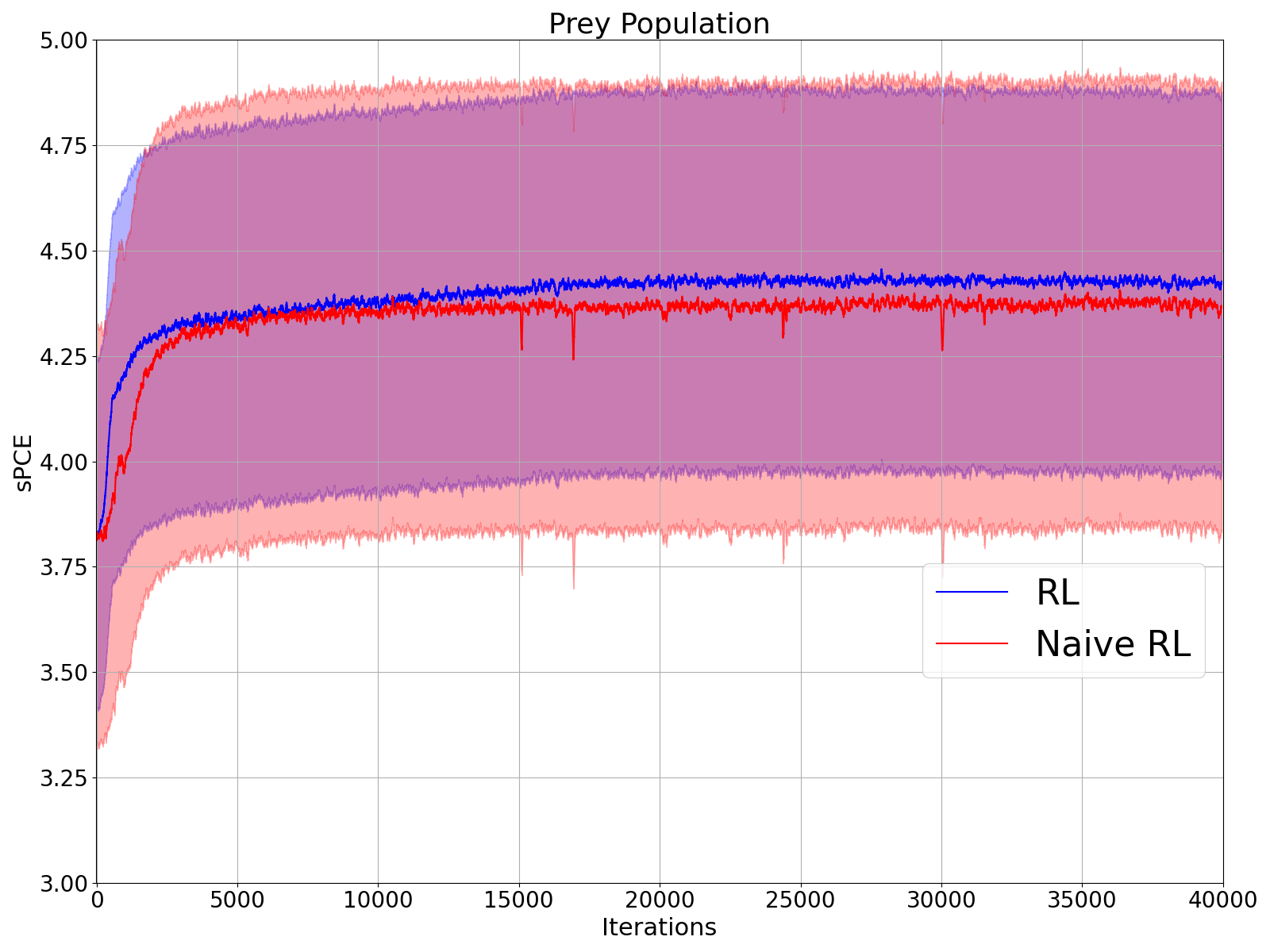}}
\vskip -0.2in
\end{figure}

\end{document}